\documentclass{article}

\usepackage[final]{neurips_data_2023}

\usepackage[utf8]{inputenc} 
\usepackage[T1]{fontenc}    
\usepackage{hyperref}       
\usepackage{url}            
\usepackage{booktabs}       
\usepackage{amsfonts}       
\usepackage{nicefrac}       
\usepackage{microtype}      
\usepackage{xcolor}         
\usepackage{adjustbox}
\usepackage{custom}
\usepackage{cleveref}

\usepackage{subcaption}
\usepackage{wrapfig}
\usepackage{lipsum}
\usepackage{listings}

\usepackage{amsmath}
\usepackage{amssymb}
\usepackage{mathtools}
\usepackage{amsthm}
\usepackage{bbm}
\usepackage{algpseudocode}
\usepackage{setspace}

\usepackage{color}
\definecolor{deepblue}{rgb}{0,0,0.5}
\definecolor{deepred}{rgb}{0.6,0,0}
\definecolor{deepgreen}{rgb}{0,0.5,0}

\newcommand\pythonstyle{\lstset{
basicstyle=\ttfamily\footnotesize,
language=Python,
morekeywords={OfflineNetHackChallengeWrapper, NetHackChallenge, get_dataset},              
keywordstyle=\color{deepblue},
emph={MyClass,__init__},          
emphstyle=\color{deepred},    
stringstyle=\color{deepgreen},
frame=single,                         
showstringspaces=false
}}

\lstnewenvironment{python}[1][]
{
\pythonstyle
\lstset{#1}
}
{}

\newcommand\pythoninline[1]{{\pythonstyle\lstinline!#1!}}

\title{Katakomba: \\Tools and Benchmarks for Data-Driven NetHack}

\makeatletter
\newcommand{\printfnsymbol}[1]{%
  \textsuperscript{\@fnsymbol{#1}}%
}
\makeatother

\author{%
  Vladislav Kurenkov\thanks{Contributed equally.} \\
  Tinkoff\\
  \texttt{v.kurenkov@tinkoff.ai} \\
  \And
  Alexander Nikulin\printfnsymbol{1} \\
  Tinkoff \\
  \texttt{a.p.nikulin@tinkoff.ai} \\
  \AND
  Denis Tarasov \\
  Tinkoff \\
  \texttt{den.tarasov@tinkoff.ai} \\
  \And
  Sergey Kolesnikov \\
  Tinkoff \\
  \texttt{s.s.kolesnikov@tinkoff.ai} \\
}

\begin{document}

\maketitle

\begin{abstract}
NetHack is known as the frontier of reinforcement learning research where learning-based methods still need to catch up to rule-based solutions. One of the promising directions for a breakthrough is using pre-collected datasets similar to recent developments in robotics, recommender systems, and more under the umbrella of offline reinforcement learning (ORL). Recently, a large-scale NetHack dataset was released; while it was a necessary step forward, it has yet to gain wide adoption in the ORL community. In this work, we argue that there are three major obstacles for adoption: resource-wise, implementation-wise, and benchmark-wise. To address them, we develop an open-source library\footnote{Source code: \url{https://github.com/corl-team/katakomba}} that provides workflow fundamentals familiar to the ORL community: pre-defined D4RL-style tasks, uncluttered baseline implementations, and reliable evaluation tools with accompanying configs and logs synced to the cloud.
\end{abstract}

\section{Introduction}

Reinforcement learning \citep{arulkumaran2017brief, levine2020offline} led to remarkable progress in the decision-making problems in recent years: robotics \citep{smithWalkParkLearning2022, kumar2021a}, recommender systems \citep{Chen2022OffPolicyAF}, traffic control \citep{chen2020toward}, energy management \citep{yu2019deep}, combinatorial optimization \citep{mazyavkina2021reinforcement}, and videogames \citep{baker2022video, hafner2023mastering}. Many of those are considered solved or close to solved problems, and the arguably next untapped frontier for RL algorithms is the NetHack \citep{kuttler2020nethack}.

NetHack\footnote{For a brief introduction to the game, we recommend excellent \href{https://nethackwiki.com/wiki/Main_Page}{game wiki}, as well as the original publication by \citet{kuttler2020nethack}, which introduced the NetHack environment.} is considered one of the most challenging games for humans, even more to learning agents. This game requires strong generalization capabilities, as the levels are procedurally generated with each reset, and extraordinary memory capacity, as the episodes may last for 100k steps requiring to remember what happened a thousand steps before to act optimally. Moreover, the high level of uncertainty and its dependence on the initial character configuration further hardens the problem. Indeed, at the moment, the best agent, AutoAscend \citep{hambro2022insights}, is essentially rule-based, and yet reinforcement learning agents cannot even come close in terms of the game score or levels reached. While there were various attempts to advance online reinforcement learning agents, this task seems out of reach for them. 

Recently, a promising direction was outlined -- re-using large-scale datasets of human/bot demonstrations to bootstrap RL agents, either in the pre-training phase or in the fine-tuning stage \citep{hambrodungeons, kumar2022offline}. This research direction holds great promise to advance the community's progress in solving the NetHack and attacking the weak sides of the RL agents. While the proposed tools and datasets are necessary, their implementations, support, and ease of adoption still need to be improved. What is essential, the use of NetHack promises the democratization of research efforts, providing way more efficient tools compared to both StarCraft \citep{vinyals2019grandmaster} and Minecraft \citep{guss2019minerl} learning environments. However, while the open-sourced datasets are a significant first step, the provided tools are very far from being readily adopted by other researchers (\Cref{sec:adoption}).

In this work, we aim to address this gap and provide a set of instruments that are conducive to future research endeavors, smooth the learning curve, and define a precise set of tasks to be evaluated against when training offline RL agents on the NetHack dataset, where contributions are as follows:

\begin{itemize}
    \item \textbf{D4RL-Style Benchmark} A set of small-scale datasets based on the large-scale data source \citep{hambrodungeons} for faster experimentation, including several data loaders for different speed-memory tradeoffs with a familiar interface to the ORL community alike \citet{d4rl}.
    \item \textbf{Clean Recurrent Offline RL Baselines} Straightforward implementations of popular offline RL baselines augmented with recurrence: BC \citep{michie1990cognitive}, CQL \citep{kumar2020conservative}, AWAC \citep{nair2020accelerating}, REM \citep{agarwal2020optimistic}, IQL \citep{kostrikov2021offline}. Each implementation is separated into single files akin to \citet{huang2021cleanrl, tarasov2022corl} to smooth the learning curve.
    \item \textbf{Evaluation Guideline} We use reliable evaluation tools for comparing offline RL agents in the natural presence of high variability. For this, we employ recent RLiable tools \citep{agarwal2021deep} that avoid comparing against mean values (which are especially deceptive in the case of the NetHack environment). In order to facilitate the comparisons, we also release the raw results we obtained in our experiments so that future work may re-use them for a fair comparison.
    \item \textbf{Open-Sourced Training Logs and Configs} Moreover, we keep a public track of our experiments: including configs, implementations, and learning dynamics using Weights\&Biases \citep{wandb}, so that it is transparent for inspection and reproduction.
\end{itemize}

We believe our work is a logical next step for advancing reinforcement learning in the NetHack environment, providing a conducive set of tools, algorithms, and utilities for experimenting with offline RL algorithms. Moreover, as we plan to support the benchmark, we intend to extend it with offline-to-online tools and benchmarks. By no means this work serves as a converged fixed point. We view this as a starting step in building reliable tools for solving NetHack with prior data and commit to further support and development.

\section{What Hinders the Adoption of Data-Driven NetHack?}
\label{sec:adoption}

In this section, we describe the main obstacles and motivate our work. We outline our experience with the recently released dataset \citep{hambrodungeons}, accompanying tools and divide the main issues into three categories: implementational, resources, and benchmark-wise. These are needed to illustrate that the initial release of the dataset is a significant step. However, it is only the first step, and further development is needed, which we will address in further sections.

\paragraph{Implementation-wise} First, when trying to reproduce the experiments, we found that installing the released tools is error-prone: where the provided Dockerfile was not operable and required fixing the moolib library with a CMake modification, which is probably undesirable for broader adoption of practitioners. Second, when we attempted to reproduce the scores reported in \citet{hambrodungeons}, the only configuration file provided was for the IQL algorithm\footnote{Consider line 38 for an example of hyperparameters ambiguity: \url{https://github.com/dungeonsdatasubmission/dungeonsdata-neurips2022/blob/main/experiment_code/hackrl/dqn_ttyrec_config.yaml}. When changing the crop dimensions, the implementation crashes.} and its modification with hyperparameters for other algorithms from the paper required additional effort to assign proper values. Consequently, the reproduction did not result near what was reported in the paper (note that a similar pattern was observed in \citet{piterbarg2023nethack}). While we do not claim that the reproduction is impossible, we could not find the proper fix in a reasonable amount of time and consider this as another obstacle in adoption -- practitioners should either have access to the original training logs or be able to replicate the results reliably.

Beyond the replication frictions, another issue lies in the design of implementations. Offline and offline-to-online settings are interleaved within just one file (1500 lines of code), where the algorithms are realized using the framework for distributed reinforcement learning not typical for the ORL community \citep{moolib2022}. Consequently, understanding the underlying logic is hard, and some performance-related implementation tricks can be hard to spot. For example, q-learning-based algorithms do not use the last element of the sequence, i.e., every $sequence\_length$ game tuple is not used for training (but only for bootstrapping). While all of these may seem like minor issues, their sum brings a significant burden upon practitioners interested in the data-driven NetHack research.

\paragraph{Resource-wise}
The great addition to the released datasets was an accompanying data loader that operates over compressed game episodes (we refer to it as TTYRec throughout the text). While the \citet{hambrodungeons} demonstrated that the proposed loader could operate at approximately 5 hours per epoch, we observed a slower rate when adapting it for recurrent algorithms. The main issue lies in the underlying format where the data is fetched the following way -- $(s_{t}, a_{t}, r_{t-1})$\footnote{Notation standard for the RL community, where $s$ is a state, $a$ is an action, and $r$ is a reward.}. While this may seem non-significant, it results in the need to fix the tuples sequence to obtain a suitable format $(s_{t}, a_{t}, s_{t+1}, r_{t})$. To better demonstrate the impact of this problem, consider \Cref{tab:loading_time}, where we test the original data loader and the one with the fix, observing a significant increase in the loading time as both batch size and sequence length increase. While this can be avoided with the original loader by simply discarding each $sequence\_length$ element (as done in the original work), the problem persists as the original implementations do not work without reward shaping with potentials requiring a scan over the sequence, not to mention that such data rejection is not justified beyond performance reasons. As an additional but important issue to resource-constrained practitioners, one must first download the entire 90GB AA dataset, even if they aim to work on a subset of data.

\vskip -0.1in
\begin{table}[h]
    \caption{Loading times for different storage formats averaged over 500 iterations. The right part of the table, with HDF5 columns, depicts the loading time for datasets repacked within Katakomba. Note that the transitions are in the proper format for them by design.}
    \label{tab:loading_time}
    \begin{center}
    \begin{small}
        \begin{adjustbox}{max width=\columnwidth}
		\begin{tabular}{l|cc|ccc}
            \toprule
            \textbf{Variants} & \textbf{TTYRec} & \textbf{TTYRec, Proper Tuples} & \textbf{HDF5 (Memmap)} & \textbf{HDF5 (RAM)} & \textbf{HDF5 (Compressed)}  \\
            \midrule
             batch\_size=64, seq\_len=16            & 15ms & 17ms & 2ms & 1ms & 516ms \\
             batch\_size=256, seq\_len=32            & 74ms & 132ms & 12ms & 9ms & 2.39s \\
             batch\_size=256, seq\_len=64            & 255ms & 372ms & 18ms & 14ms & 2.42s \\
            \bottomrule
		\end{tabular}
        \end{adjustbox}
    \end{small}
    \end{center}
    \vskip -0.1in
\end{table}

\paragraph{Benchmark-wise} Arguably, one of the driving forces in Deep Learning and Reinforcement Learning, in general, is a well-defined set of tasks. To demonstrate how the proposed large-scale dataset could be utilized for tasks definition, \citet{hambrodungeons} described two settings: learning on the whole dataset for all role-race-alignment-sex combinations and learning on the subset of data for the Monk-Human-Neutral setting. While this is a good entry point and could be of great use to practitioners interested in large-scale data regimes, there was no detailed proposal on how one should further define tasks which is indeed an open question in the NetHack learning community \citep{kuttler2020nethack}. Moreover, the original raw large-scale dataset requires practitioners to manually define SQL queries for extracting the data of interest, which is flexible but could be an overkill.

More importantly, the proposed comparison metrics were mean and median statistics of average episode returns over training seeds. One may argue that the median is a robust statistic. However, in this case, it is a median of \textit{average} episode returns over training seeds and not of the whole set of evaluation episodes. It is well known in the RL community \citep{agarwal2021deep} that those are not reliable due to the highly-variable nature of RL agents. As of the NetHack, this further amplifies by the extremely-variable nature of the game itself as both demonstrated in \citet{kuttler2020nethack, hambrodungeons}. Therefore, to reduce the noise in the progress of RL agents in this domain, one would need a different evaluation toolset that is better suited for it.


\section{Katakomba}





Given the issues described in the previous section, we are ready to present Katakomba -- a set of tools, benchmarks, and memory-based offline RL baselines. In this section, we gradually introduce the components of the library with accompanying code examples to better demonstrate the ease of use. All the numbers and performance metrics discussed in the text were measured on the same set of hardware: 14CPU, 128GB RAM, NVMe, 1xA100; for more details, please, see \cref{appendix:res-stats}.

\subsection{Benchmark}
\begin{python}[caption={Usage example: Tasks are defined via the character field that is then used by the offline wrapper for dataset loading and score normalization.}]
from katakomba.env import NetHackChallenge
from katakomba.env import OfflineNetHackChallengeWrapper

# The task is specified using the character field
env = NetHackChallenge(
    character="mon-hum-neu", 
    observation_keys=["tty_chars", "tty_colors", "tty_cursor"]
)

# A convenient wrapper that provides interfaces for
# dataset loading and score normalization
env = OfflineNetHackChallengeWrapper(env)

\end{python}
\vspace{-0.1in}
\begin{table}[h]
    \caption{Katakomba tasks. We split the large-scale AutoAscend dataset into 38 problems dividing them into three categories. The splits are justified by the early-game nature of the AutoAscend data, where roles have a higher impact on the gameplay, then races, and then alignments. Note that the whole benchmark covers all possible role-race-alignment combinations.}
    \label{tab:tasks}
    \begin{center}
    \begin{small}
        \begin{adjustbox}{max width=\columnwidth}
		\begin{tabular}{l|cccccc}
            \toprule
            \textbf{Tasks} & \textbf{\# Transitions} & \textbf{Median Turns} & \textbf{Median Score} & \textbf{Median Deathlvl} & \textbf{Size (GB)} & \textbf{Compressed Size (GB)} \\
            \midrule
            \textbf{Base (Role-Centric)} & - & - & - & - & - & - \\
            \midrule
             \underline{arc}-hum-neu             & 24527163 & 32858.0 & 4802.5  & 2.0 & 94.5  & 1.3 \\
             \underline{bar}-hum-neu             & 26266771 & 35716.0 & 11964.0 & 4.0 & 101.1 & 1.7 \\
             \underline{cav}-hum-neu             & 21674680 & 30361.0 & 8152.0  & 4.0 & 83.5  & 1.3 \\
             \underline{hea}-hum-neu             & 14473997 & 18051.0 & 2043.0  & 1.0 & 55.7  & 0.8 \\
             \underline{kni}-hum-law             & 22287283 & 28246.0 & 6305.0  & 3.0 & 85.8  & 1.5 \\
             \underline{mon}-hum-neu             & 33741542 & 42400.0 & 11356.0 & 4.0 & 129.9 & 2.1 \\
             \underline{pri}-hum-neu             & 18376473 & 26796.5 & 5366.5  & 2.0 & 70.8  & 1.1 \\
             \underline{ran}-hum-neu             & 17625493 & 25354.0 & 6168.0  & 2.0 & 67.9  & 1.0 \\
             \underline{rog}-hum-cha             & 14284927 & 19334.0 & 3005.5  & 1.0 & 55.0  & 0.8 \\
             \underline{sam}-hum-law             & 22422537 & 32951.0 & 7850.0  & 4.0 & 86.3  & 1.3 \\
             \underline{tou}-hum-neu             & 13376498 & 17955.5 & 2554.5  & 1.0 & 51.5  & 0.8 \\
             \underline{val}-hum-neu             & 27784788 & 35250.0 & 11402.5 & 4.0 & 107.0 & 1.8 \\
             \underline{wiz}-hum-neu             & 14343449 & 19808.5 & 3132.5  & 1.0 & 55.2  & 0.8 \\
             \midrule
             \textbf{Extended (Race-Centric)} & - & - & - & - & - & - \\
             \midrule
             pri-\underline{elf}-cha             & 18796560 & 26909.5 & 4718.5  & 2.0 & 72.4  & 1.1 \\
             ran-\underline{elf}-cha             & 18238686 & 26607.0 & 7583.0  & 4.0 & 70.2  & 1.1 \\
             wiz-\underline{elf}-cha             & 15277820 & 19512.0 & 2988.5  & 1.0 & 58.8  & 0.9 \\
             arc-\underline{dwa}-law             & 25100788 & 34669.0 & 4026.0  & 1.0 & 96.7  & 1.5 \\
             cav-\underline{dwa}-law             & 22871890 & 32261.0 & 7158.0  & 3.0 & 88.1  & 1.5 \\
             val-\underline{dwa}-law             & 32787658 & 33973.0 & 8652.5  & 3.0 & 126.6 & 2.5 \\
             arc-\underline{gno}-neu             & 24144048 & 34432.0 & 4077.5  & 1.0 & 93.0  & 1.4 \\
             cav-\underline{gno}-neu             & 21624779 & 29860.0 & 6446.0  & 3.0 & 83.3  & 1.4 \\
             hea-\underline{gno}-neu             & 14884704 & 18518.0 & 1980.5  & 1.0 & 57.3  & 0.9 \\
             ran-\underline{gno}-neu             & 17571659 & 25970.0 & 5326.0  & 2.0 & 67.7  & 1.1 \\
             wiz-\underline{gno}-neu             & 14193637 & 19206.0 & 2736.0  & 1.0 & 54.7  & 0.9 \\
             bar-\underline{orc}-cha             & 27826356 & 39291.0 & 10499.0 & 4.0 & 107.2 & 1.8 \\
             ran-\underline{orc}-cha             & 18127448 & 26707.0 & 5460.0  & 2.0 & 69.8  & 1.1 \\
             rog-\underline{orc}-cha             & 16674806 & 22351.0 & 3103.0  & 1.0 & 64.2  & 1.0 \\
             wiz-\underline{orc}-cha             & 15994150 & 22570.5 & 3241.5  & 1.0 & 61.6  & 1.0 \\
              \midrule
              \textbf{Complete (Alignment-Centric)} & - & - & - & - & - & - \\
             \midrule
             arc-hum-\underline{law}             & 23422383 & 31446.0 & 4188.0  & 1.0 & 90.2  & 1.3 \\
             cav-hum-\underline{law}             & 22328494 & 31039.0 & 8174.0  & 4.0 & 86.0  & 1.3 \\
             mon-hum-\underline{law}             & 30782317 & 39647.0 & 10855.0 & 4.0 & 118.5 & 1.9 \\
             pri-hum-\underline{law}             & 18298816 & 27192.0 & 4833.0  & 1.0 & 70.5  & 1.1 \\
             val-hum-\underline{law}             & 30171035 & 34570.5 & 9707.0  & 4.0 & 116.2 & 2.1 \\
             bar-hum-\underline{cha}             & 25362111 & 35925.0 & 12574.0 & 5.0 & 97.7  & 1.6 \\
             mon-hum-\underline{cha}             & 33662420 & 41730.5 & 11418.0 & 4.0 & 129.6 & 2.1 \\
             pri-hum-\underline{cha}             & 18667816 & 28204.5 & 5847.0  & 2.0 & 71.9  & 1.1 \\
             ran-hum-\underline{cha}             & 16999630 & 24698.5 & 6236.0  & 2.0 & 65.6  & 1.0 \\
             wiz-hum-\underline{cha}             & 14635591 & 20257.0 & 3294.0  & 1.0 & 56.4  & 0.9 \\
            \bottomrule
		\end{tabular}
        \end{adjustbox}
    \end{small}
    \end{center}
    \vskip -0.1in
\end{table}

\paragraph{Decomposition} 
In our benchmark, we re-use the dataset collected by the AutoAscend bot \citep{hambrodungeons, hambro2022insights}. While this bot is highly-capable, it still is considered an early-game contender because it can not descend further than two levels in half of the episodes. As the dataset becomes an early game bridgehead, we divide the tasks based on the character configurations: role, race, and alignment. In the NetHack community, these are known to be the most important (and even having a dramatic effect), requiring to utilize varying abilities of each role or race\footnote{\url{https://nethackwiki.com/wiki/Player}}. Overall, in opposition to merging all combinations, this decomposition allows more focus on the characters’ gameplay and the size reduction one needs to download for both playing-around and committed research, as we will describe further.

\paragraph{Categories}
This results in 38 datasets in total. However, it may not be possible for researchers to examine each of them as the training times can be an obstacle. To this end, we further divide the tasks into three categories: \underline{Base}, \underline{Extended}, and \underline{Complete}. We separate each category based on the wisdom of the NetHack community, i.e., that roles have a more substantial effect on the gameplay than race, and race has more effect than alignment. The datasets and categories are listed in \cref{tab:tasks}. \underline{Base} tasks consist of all possible roles for the human race; we choose a neutral alignment where possible; otherwise, we pick the alternative one (for humans, if there is no neutral alignment, there is only one alternative). We include all other role-race combinations for \underline{Extended} tasks. For \underline{Complete}, we add tasks that were not included in the Base and Extended categories. Note that these three categories combined cover all of the allowed role-race-alignment combinations.

\paragraph{Data Selection} As demonstrated in the previous section, the original TTYRec dataset is actually slower than expected when it is used for learning agents. Moreover, one may be unable to download the 90GB dataset. Therefore, we take a different path by subsampling and repacking the original dataset task-wise, averaging 680 episodes and 1.3GB per task. The subsampling procedure is stratified by the game score (for more details, please see the script\footnote{\url{https://github.com/corl-team/katakomba/blob/main/scripts/generate_small_dataset.py}}). This allows for more versatility: one can download the needed datasets on demand as in D4RL \citep{d4rl}; furthermore, this permits us to address the rolling issue as we repacked the transition tuples in the way suitable to typical ORL pipeline as a part of the release. To ensure the reliable accessibility of the data, we host it on the HuggingFace Hub \citep{Engstrom2020ImplementationMI}.

\subsection{Data Loaders for Speed-Memory Tradeoff}
As we outlined in \Cref{sec:adoption}, while the original large-scale dataset is well-compressed, its iteration speed integrated with proper sequential loaders is still an obstacle for fast experimentation loop. Moreover, it requires at least one full download for the entire large-scale dataset, which may not be suitable for resource-constrained scenarios. In Katakomba, we address this by providing three different loaders trading-off memory and speed allowed by the deliberate task division. In each of them, the compressed task's dataset is automatically downloaded and decompressed if one aims for speed. Moreover, we provide a Python interface for an automatic clean-up if one does not want to store the decompressed dataset beyond the code execution.

\paragraph{HDF5, In-Memory (Decompressed, RAM)} This is the fastest format that relies on the decompression of the dataset into the RAM, which may require from 51GB to 129GB depending on the dataset listed in \Cref{tab:tasks}. This option is quite demanding regarding the memory but comes with the advantage of the fastest data access (see \Cref{tab:loading_time}).

\paragraph{HDF5, Memory Map (Decompressed, Disk)} This option is a middle-ground that allows the datasets to be efficiently utilized without requiring large RAM. When one specifies this preference, the dataset will be decompressed on one's hard drive, and the consequent reads will be conducted from it. The advantage of this approach compared to the loader from \citet{hambrodungeons} is that there is no need to decompress on the fly, and this is done precisely once at the start of the training process (taking from three to ten minutes on average).

\begin{python}[caption={Usage example: Katakomba provides different loaders that can be chosen depending on one's speed-memory tradeoff.}]
# Decompress the dataset into RAM
dataset = env.get_dataset(mode="in_memory")

# Decompression on-read
dataset = env.get_dataset(mode="compressed")

# The original dataloader introduced in Hambro et al, 2022
dataset = env.get_dataset(scale="big")

# Decompress the dataset on disk
dataset = env.get_dataset(mode="memmap")

# If you want to delete the decompressed dataset
# This will not affect the compressed version
dataset.close()

\end{python}

\paragraph{HDF5, In-Disk (Compressed, Disk)} This mode is the most cheap one but slow. Essentially, the dataset is read on from disk and decompressed on-the-fly. We found this useful for debugging purposes where one does not need the whole training process to be run.

\paragraph{TTYRec, In-Disk (Compressed, Disk)} In case one finds the original approach to data loading more suitable, we also provide a convenient interface that wraps the source large-scale dataset and loader. One can also set it up in more detail using keyword arguments from the original TTYRec data loader. However, this option comes with the downsides described in \Cref{sec:adoption}, i.e., slower iteration speed and the need to download the 90GB dataset at least once.

In addition to the dataset interfaces, we also provide an implementation of a replay buffer suitable for sequential offline RL algorithms for bootstrapping practitioners with ready-to-go tools.

\subsection{Evaluation Methodology under High Variability}
\label{subsec:evaluation}

In \citet{hambrodungeons}, authors used an average episode return across seeds when comparing baselines. While this is a standard practice in the RL and ORL communities, it was recently shown to be unreliable \citep{agarwal2021deep} as the algorithms are known to be highly variable in performance. This problem amplifies even more for NetHack, where the score distribution is typically quite wide for humans and bots \citep{kuttler2020nethack}. 

To address this, we argue that the evaluation toolbox from \citet{agarwal2021deep} is more appropriate and suggest using it when comparing NetHack learning agents. We use these tools for two dimensions: in-game score and death level. The first dimension corresponds to what one typically optimizes with ORL agents but is considered a proxy metric \citep{kuttler2020nethack}. While the latter lower bounds the early-game progress and is more indicative of the game completion. 

Furthermore, similar to \citet{d4rl}, we also suggest reporting normalized scores to capture how far one is from the AutoAscend bot. We use mean scores per dataset as a normalization factor and rescale to [0, 100] range after normalization, similar to D4RL \citep{d4rl}. This functionality is also provided as a part of the offline wrapper for the NetHackChallenge environment. Please refer to \Cref{appendix:res-stats} for precise values.

\section{Benchmarking Recurrent Offline RL Algorithms}
\label{sec:benchmark}
\begin{figure}[ht]
\centering
\captionsetup{justification=centering}
     \centering
         \begin{subfigure}[b]{0.98\textwidth}
         \centering
         \includegraphics[width=1.0\textwidth]{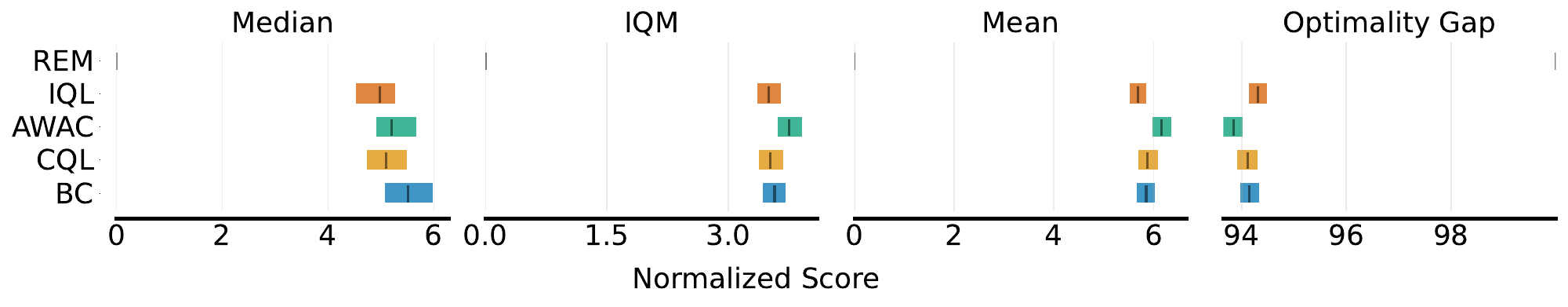}
         \caption{Bootstrapped point estimates. It can be seen that except REM, which failed, all algorithms perform about the same across all metrics and are generally far behind the
         AutoAscend algorithm, which mean scores per dataset were used for normalization.}
         \label{fig:ns:point-estimates}
        \end{subfigure}
        \begin{subfigure}[b]{0.49\textwidth}
         \centering
         \includegraphics[width=1.0\textwidth]{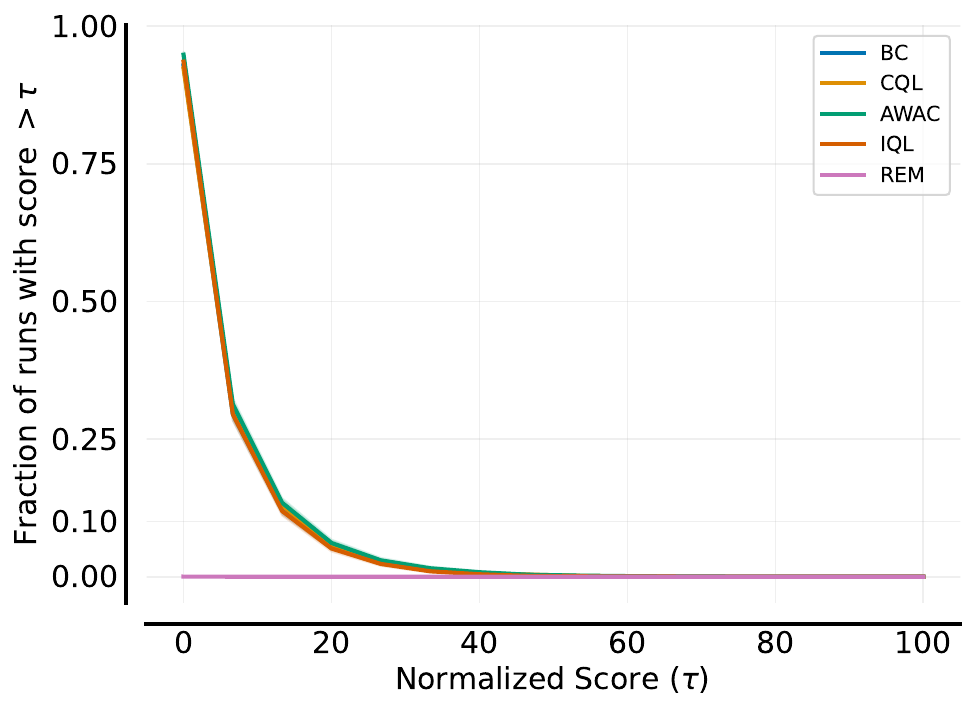}
         \caption{Performance profiles. Notably, around 5\% of episodes can surpass the normalized score of 20.}
         \label{fig:ns:performance-profiles}
        \end{subfigure}
        \begin{subfigure}[b]{0.49\textwidth}
         \centering
         \includegraphics[width=1.0\textwidth]{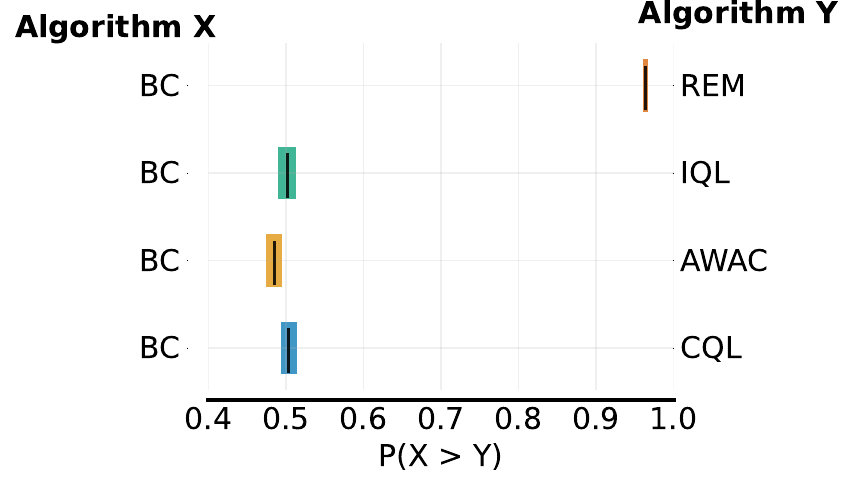}
         \caption{Probability of improvement of BC to other algorithms.}
         \label{fig:ns:probability-improvement}
        \end{subfigure}
        \caption{{Normalized performance under the Katakomba benchmark for all 38 datasets. Each algorithm was run for three seeds and evaluated over 50 episodes resulting in 5700 points for constructing these graphs. As one can see, there is not much improvement beyond naive behavioral cloning.}}
        \label{fig:ns}
\end{figure}

\paragraph{Algorithms} For our benchmark, we rely on the following set of algorithms: Behavioral Cloning (BC) \citep{michie1990cognitive}, Implicit Q-Learning (IQL) \citep{kostrikov2021offline}, Conservative Q-Learning (CQL) 
\citep{kumar2020conservative}, Randomized Ensemble Mixture (REM)  \citep{agarwal2020optimistic}, and Advantage-Weighted Actor-Critic (AWAC) \citep{nair2020accelerating}. These are known as either the most competitive in the continuous control setting \citep{tarasov2022corl} or were shown to be competitive in the discrete control \citep{agarwal2020optimistic, Kumar2022OfflineQO}. Similar to \citet{hambro2022insights, hambrodungeons}, we build upon Chaotic-Dwarven-GPT-5 architecture that converts the tty observation into an image and then feeds it into the CNN layers followed by the LSTM \citep{lstm}. Ultimately, we test five common ORL algorithms augmented with recurrence. To the best of our knowledge, there are no other open-sourced alternatives beyond the \citet{hambro2022insights} that also do not implement the AWAC algorithm.

\paragraph{Experimental Setup} We train every algorithm for 500k gradient steps, resulting in around 20 epochs per dataset. We report the scores of the last trained policy over 50 evaluation episodes as standard in the ORL community. Important to highlight that while this amount may seem small for NetHack, this number is adequate when used in conjunction with stratified datasets and RLiable evaluation tools due to the underlying bootstrapping mechanism. For specific hyperparameters used, please either see \Cref{appendix:hyperparams} or the configuration files provided in the code repository.

\paragraph{Replicability and Reproducibility Statement} To ensure that our experiments' results are replicable and can easily be reproduced and inspected, we rely on the Weights\&Biases \citep{wandb}. In the provided code repository, one can find all the logs, including configs, dependencies, code snapshots, hyperparameters, system variables, hardware specifications, and more. Moreover, to reduce the efforts of interested parties required for inspection, we structurize the results using the Weights\&Biases public reports.

\begin{figure}[ht]
\centering
\captionsetup{justification=centering}
     \centering
         \begin{subfigure}[b]{0.98\textwidth}
         \centering
         \includegraphics[width=1.0\textwidth]{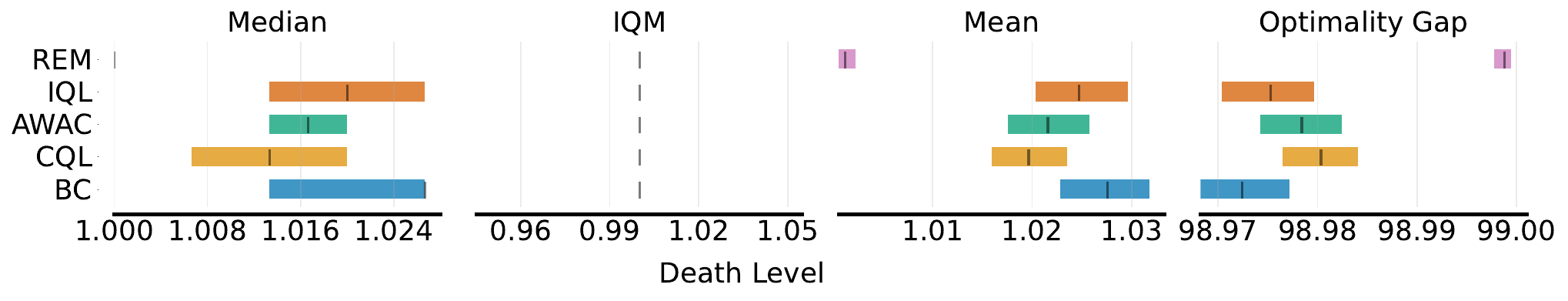}
         \caption{Bootstrapped point estimates.}
         \label{fig:dthlvl:point-estimates}
        \end{subfigure}
        \begin{subfigure}[b]{0.49\textwidth}
         \centering
         \includegraphics[width=1.0\textwidth]{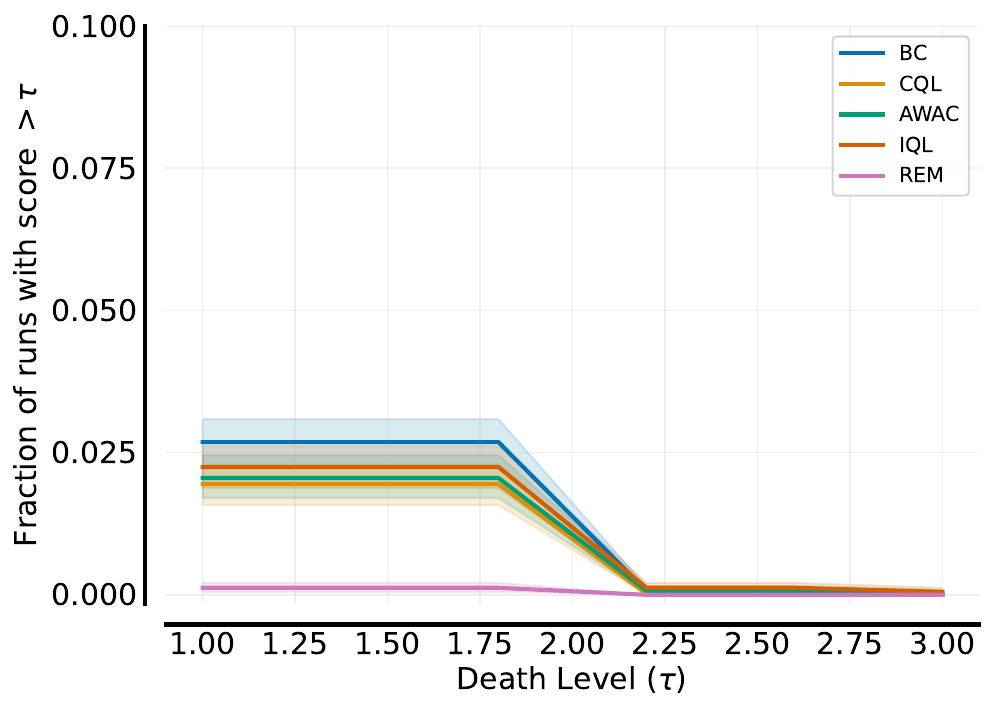}
         \caption{Performance profiles. Notably, only 2.5\% of episodes progress to the second level.}
         \label{fig:dthlvl:performance-profiles}
        \end{subfigure}
        \begin{subfigure}[b]{0.49\textwidth}
         \centering
         \includegraphics[width=1.0\textwidth]{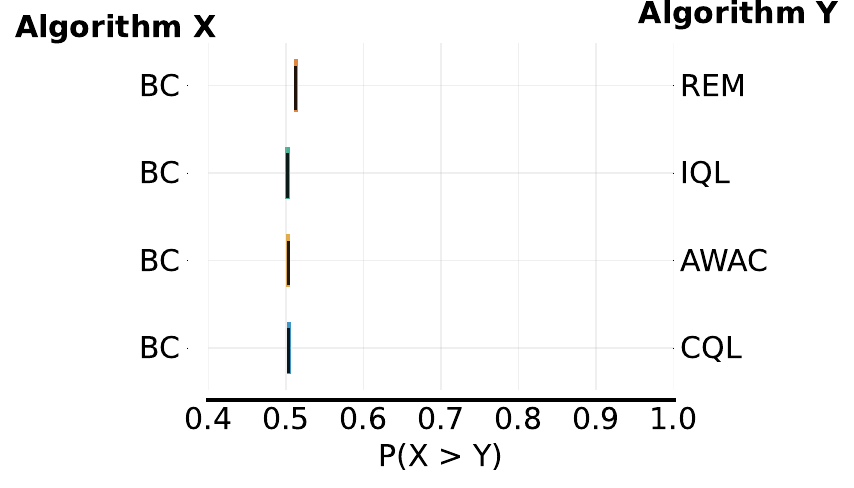}
         \caption{Probability of improvement of BC to other algorithms.}
         \label{fig:dthlvl:probability-improvement}
        \end{subfigure}
        \caption{{Death level under the Katakomba benchmark for all 38 datasets. Each algorithm was run for three seeds and evaluated over 50 episodes resulting in 570 points for constructing these graphs. As one can clearly see, there is not much improvement beyond the naive behavioral cloning.}}
        \label{fig:dthlvl}
\end{figure}

\paragraph{Results} The outcomes are twofold. First, the results achieved are similar to the already observed by the \citet{piterbarg2023nethack} and \citet{hambrodungeons}, but on a larger number of offline RL algorithms tested. As shown in \Cref{fig:ns} and \Cref{fig:dthlvl}, all algorithms were unable to replicate the scores of the AutoAscend bot, reaching normalized scores below 6.0 on average and not progressing beyond the first level on the majority of training runs. Notably, only 5\% of the episodes resulted in a normalized score of around 20.0 (\Cref{fig:ns:performance-profiles}). Moreover, REM has not been able to achieve even the non-zero normalized score. 
Second, perhaps surprisingly, the only algorithm that does not rely in any way on policy constraints is also the only algorithm that completely failed. This, and also the high correlation in the performance profiles (\Cref{fig:ns:performance-profiles}), gives us a hint that all other methods showing non-zero results actually rely primarily on behavioral cloning in one form or another, such as advantage weighted regression in IQL and AWAC or KL-divergence as in CQL. Indeed, the most successful hyperparameters in our experiments proved to be those that significantly increase the weight of losses that encourage similarity to the behavioral policy (see \Cref{app:cql-alphas} in the \Cref{appendix:hyperparams}). Furthermore, as shown in the \Cref{fig:ns:probability-improvement} and \Cref{fig:ns:probability-improvement} Behavioral Cloning algorithm is not worse than all the other more sophisticated offline RL algorithms. Thus, NetHack remains a major challenge for offline RL algorithms, and Katakomba can serve as a starting point and testbed for offline RL research in this direction. For graphs stratified by the Base, Extended, and Complete categories, see \cref{appendix:results-stratified}. 


\section{Related Work}

\paragraph{Offline RL} In recent years, there was considerable interest of the reinforcement learning community in the offline setting, which resulted in numerous and diverse algorithms specifically tailored for this setup \citep{nair2020accelerating, kumar2020conservative, kostrikov2021offline, chen2021decision, fujimoto2021minimalist, an2021uncertainty, nikulin2023anti, tarasov2023revisiting}. The core idea behind most of them is to ensure the resulting policy stays close to the behavioral one. This could be achieved via different ways: penalizing value function \citep{kumar2020conservative, an2021uncertainty, nikulin2023anti}, constraining the policy outputs \citep{fujimoto2021minimalist, tarasov2023revisiting}, or even training directly in the conservative latent space \citep{PLAS_corl2020, chen2022latent, akimov2022let}. Due to the benchmark-centricity of the RL field, most of the proposed ORL algorithms are for continuous control with a few exceptions \citep{agarwal2020optimistic, kumar2020conservative, Kumar2022OfflineQO}. The de-facto standard benchmark is the D4RL \citep{d4rl}, which provides a suite of datasets focused on continuous control with proprioceptive states under different regimes, such as sparse-reward or low-data regimes. Also, few benchmarks move the focus from proprioceptive states to images or other more complex entities \citep{qin2022neorl, lu2022challenges, agarwal2020optimistic}.

\paragraph{RL for NetHack} NetHack as a testbed for RL agents was introduced in \citet{kuttler2020nethack}. To further advance the RL agents in this domain, the NetHack Challenge Competition (NHC) was held \citep{hambro2022insights} that resulted in two of the most performant agents -- learning-based Chaotic-Dwarven-GPT-5, and a rule-based AutoAscend (AA). Notably, the latter outperformed learning-based agents by a wide margin. Consequently, this solution was used to collect the large-scale NetHack Learning Dataset \citep{hambrodungeons}. The closest concurrent work is by \citet{piterbarg2023nethack} -- the authors released another AA dataset but accompanied with the hierarchical labels, which arise due to the nature of the AutoAscend bot, demonstrating their usefulness in cloning the AA bot. However, in contrast to our work, \citet{piterbarg2023nethack} focuses on the large-scale setup similar to \citet{hambrodungeons}.

\section{Discussion, Limitations, and Future Work}
\label{discussion}
In this work, we focused on building reliable tools and benchmarks for offline RL setting using the recently released AutoAscend large-scale dataset \citep{hambrodungeons}. While this does not cover the whole spectrum of NetHack's community interests, such as offline-to-online regime or learning from human demonstrations, we believe our effort is helpful in establishing reliable and open instruments for data-driven NetHack. Moreover, our contributions could be of interest to the part of the ORL community studying discrete control, memory, and adaptivity \citep{ghosh2022offline}.

Given our results and experience with the NetHack Learning Environment, we believe fruitful future research may lie along the following directions: finding a better state-encoder, as the current one presents a bottleneck in both efficiency (rendering is expensive) and locality (only the small part of the terminal is used). Another interesting research direction would be to assess recently appeared recurrence mechanisms such as Linear Recurrent Unit \citep{orvieto2023resurrecting}, which might also speed up the training process without hurting the performance. Also, as the interest in generalization properties will appear, it would be a great addition to include more datasets that will provide metadata on the seeds used for data generation, as it will allow to assess trained agents on both seen and unseen seeds to quantify the generalization gap more systematically.

Overall, we firmly believe that NetHack provides a nice playground for investigating how to build a next generation of reinforcement learning agents using prior data that would encompass stronger generalization and memory capabilities. To this end, we plan to continuously maintain the benchmark, accompanying tools, and curate new datasets if considered useful for further advancements.

\bibliography{bib}
\bibliographystyle{iclr2023_conference}

\newpage
\section*{Checklist}

\begin{enumerate}

\item For all authors...
\begin{enumerate}
  \item Do the main claims made in the abstract and introduction accurately reflect the paper's contributions and scope?
    \answerYes{}
  \item Did you describe the limitations of your work?
    \answerYes{See \Cref{discussion}}
  \item Did you discuss any potential negative societal impacts of your work?
    \answerYes{See \Cref{appendix:ethics}}
  \item Have you read the ethics review guidelines and ensured that your paper conforms to them?
    \answerYes{See \Cref{appendix:ethics}}
\end{enumerate}

\item If you are including theoretical results...
\begin{enumerate}
  \item Did you state the full set of assumptions of all theoretical results?
    \answerNA{}
	\item Did you include complete proofs of all theoretical results?
    \answerNA{}
\end{enumerate}

\item If you ran experiments (e.g. for benchmarks)...
\begin{enumerate}
  \item Did you include the code, data, and instructions needed to reproduce the main experimental results (either in the supplemental material or as a URL)?
    \answerYes{We release our codebase, configs, and in-depth reports at \url{https://github.com/corl-team/katakomba}}
  \item Did you specify all the training details (e.g., data splits, hyperparameters, how they were chosen)?
    \answerYes{See \Cref{sec:benchmark} and \Cref{appendix:hyperparams}}
	\item Did you report error bars (e.g., with respect to the random seed after running experiments multiple times)?
    \answerYes{We extensively argued for and relied on a more robust and suitable for reinforcement learning reporting methodology based on performance profiles. For more discussion, see \Cref{subsec:evaluation}.}
	\item Did you include the total amount of compute and the type of resources used (e.g., type of GPUs, internal cluster, or cloud provider)?
    \answerYes{See \Cref{appendix:res-stats}.}
\end{enumerate}

\item If you are using existing assets (e.g., code, data, models) or curating/releasing new assets...
\begin{enumerate}
  \item If your work uses existing assets, did you cite the creators?
    \answerYes{}
  \item Did you mention the license of the assets?
    \answerYes{See \Cref{appendix:license}.}
  \item Did you include any new assets either in the supplemental material or as a URL?
    \answerYes{The links to the repacked datasets can be found in our repository: \url{https://github.com/corl-team/katakomba}.  Moreover, one can also find the scripts for generating the datasets.}
  \item Did you discuss whether and how consent was obtained from people whose data you're using/curating?
    \answerYes{See both \Cref{appendix:license,appendix:ethics}}
  \item Did you discuss whether the data you are using/curating contains personally identifiable information or offensive content?
    \answerNA{}
\end{enumerate}

\item If you used crowdsourcing or conducted research with human subjects...
\begin{enumerate}
  \item Did you include the full text of instructions given to participants and screenshots, if applicable?
    \answerNA{}
  \item Did you describe any potential participant risks, with links to Institutional Review Board (IRB) approvals, if applicable?
    \answerNA{}
  \item Did you include the estimated hourly wage paid to participants and the total amount spent on participant compensation?
    \answerNA{}
\end{enumerate}

\end{enumerate}


\clearpage
\appendix

\section{What Is Inside the Datasets?}

Every dataset is repacked into HDF5 files similar to \citet{d4rl}. The data keys are described in \Cref{tab:observation}; along to the typical $(s_{t}, a_{t}, r_{t}, d_{t})$ tuples, the metadata is also provided as the datasets' attributes with a comprehensive information about specific trajectories similar to \citet{hambrodungeons}. The re-packing script is provided at \url{https://github.com/corl-team/katakomba/tree/main/scripts/generate_small_dataset.py}.

\begin{table}[h]
    \caption{The re-packed datasets constitute of transformed data from \citet{hambrodungeons}. Dissimilar the the large scale dataset, the repacked data is now in the format familiar to the ORL practitioners. We also save the metadata for each trajectory, for a comprehensive description, please, see Appendix F in \citet{hambrodungeons}.}
    \centering
    \begin{tabular}{l c c p{7.5cm}}
         \toprule
         Name &  Type & Shape &  Description\\
         \midrule
        \texttt{tty\_chars} & \texttt{np.uint8} & [B, T, H, W] & $s_{t}$: The on-screen characters (default screen size 80 x 24).\\
        \texttt{tty\_colors} & \texttt{np.int8} & [B, T, H, W] & $s_{t}$: The on-screen colors for each character. \\
        \texttt{tty\_cursor} & \texttt{np.int16} & [B, T, 2] & $s_{t}$: The coordinates of the on-screen cursor.\\
        \texttt{actions} & \texttt{np.uint8} & [B, T] & $a_{t}$: The NLE actions the player made in response to the $s_{t}$. \\
        \texttt{rewards} & \texttt{np.int32} & [B, T] & $r_{t}$: The difference between in-game scores at states $s_{t}$ and $s_{t-1}$. Note that this was used in all implementations of the algorithms provided in \citet{hambrodungeons}. We also found that without this reward-shaping, all offline RL algorithms failed completely. \\
        \texttt{dones} & \texttt{np.uint8} & [B, T] & $d_{t}$: An indicator whether the current state is the last one in the trajectory.\\
    
        \bottomrule
    \label{tab:observation}
    \end{tabular}
\end{table}

\section{License}
\label{appendix:license}
Our codebase and repacked datasets are released under the NETHACK GENERAL PUBLIC LICENSE. The original NetHack Learning environment \citep{kuttler2020nethack} and large-scale datasets \citep{hambrodungeons} are also released under NETHACK GENERAL PUBLIC LICENSE.

\section{General Ethic Conduct and Potential Negative Societal Impact}
\label{appendix:ethics}

To the best of our knowledge, our work does not present any direct potential negative societal impact. 

As of the general ethic conduct, we believe that the most relevant issue to be discussed is the "Consent to use or share the data". Our work is largely built upon both the NetHack Learning Environment \citep{kuttler2020nethack} and the coresponding large-scale dataset \citep{hambrodungeons}, and as already described in the \Cref{appendix:license} both are distributed under the NETHACK GENERAL PUBLIC LICENSE that explicitly allows for re-usage and re-distribution.

\newpage
\section{Resources and Statistics}
\label{appendix:res-stats}

We used 64 separated computational nodes with 1xA100, 14CPU, 128GB RAM, and the NVMe as long-term storage for all our experiments. All the values reported in the paper were also obtained under this configuration. One can also find more detailed information inside the Weights\&Biases logs in the code repository.

\vspace{-0.1in}
\begin{table}[h]
    \caption{Scores used for Normalization. You can also find them at \url{https://github.com/corl-team/katakomba/blob/main/katakomba/utils/scores.py}. For other statistics, please, see \Cref{tab:tasks} in the main text.}
    \label{appendix:tab:scores}
    \begin{center}
    \begin{small}
        \begin{adjustbox}{max width=\columnwidth}
		\begin{tabular}{l|rrr}
            \toprule
            \textbf{Tasks} & \textbf{Minimum Score} & \textbf{Maximum Score} & \textbf{Mean Score} \\
            \midrule
            \textbf{Base (Role-Centric)} & - & - & - \\
            \midrule
             \underline{arc}-hum-neu             & 0.0 & 138103.0 & 6636.44 \\
             \underline{bar}-hum-neu             & 0.0 & 292342.0 & 17836.68 \\
             \underline{cav}-hum-neu             & 0.0 & 258978.0 & 12113.87 \\
             \underline{hea}-hum-neu             & 0.0 & 64337.0 & 4068.27 \\
             \underline{kni}-hum-law             & 0.0 & 419154.0 & 14137.06 \\
             \underline{mon}-hum-neu             & 0.0 & 171224.0 & 17456.05 \\
             \underline{pri}-hum-neu             & 0.0 & 114269.0 & 7732.69 \\
             \underline{ran}-hum-neu             & 0.0 & 54874.0 & 8067.99 \\
             \underline{rog}-hum-cha             & 0.0 & 68628.0 & 4818.20 \\
             \underline{sam}-hum-law             & 0.0 & 155163.0 & 11009.36 \\
             \underline{tou}-hum-neu             & 0.0 & 59484.0 & 4211.47 \\
             \underline{val}-hum-neu             & 16.0 & 313858.0 & 18624.77 \\
             \underline{wiz}-hum-neu             & 0.0 & 71709.0 & 5323.48 \\
             \midrule
             \textbf{Extended (Race-Centric)} & - & - & -  \\
             \midrule
             pri-\underline{elf}-cha             & 0.0 & 83744.0 & 7109.35 \\
             ran-\underline{elf}-cha             & 0.0 & 66690.0 & 9014.18 \\
             wiz-\underline{elf}-cha             & 0.0 & 71664.0 & 5005.16 \\
             arc-\underline{dwa}-law             & 0.0 & 83496.00 & 5445.69 \\
             cav-\underline{dwa}-law             & 0.0 & 161682.0 & 11893.48 \\
             val-\underline{dwa}-law             & 0.0 & 1136591.0 & 23473.61 \\
             arc-\underline{gno}-neu             & 0.0 & 110054.0 & 5316.57 \\
             cav-\underline{gno}-neu             & 0.0 & 142460.0 & 10083.06 \\
             hea-\underline{gno}-neu             & 0.0 & 69566.0 & 3783.93 \\
             ran-\underline{gno}-neu             & 0.0 & 58137.0 & 6965.04 \\
             wiz-\underline{gno}-neu             & 0.0 & 37376.0 & 4317.51 \\
             bar-\underline{orc}-cha             & 0.0 & 164296.0 & 17594.38 \\
             ran-\underline{orc}-cha             & 3.0 & 69244.0 & 7608.48 \\
             rog-\underline{orc}-cha             & 0.0 & 54892.0 & 4897.69 \\
             wiz-\underline{orc}-cha             & 0.0 & 40871.0 & 5016.74 \\
              \midrule
              \textbf{Complete (Alignment-Centric)} & - & - & - \\
             \midrule
             arc-hum-\underline{law}             & 2.0 & 84823.0 & 5826.35 \\
             cav-hum-\underline{law}             & 0.0 & 156966.0 & 12462.82 \\
             mon-hum-\underline{law}             & 7.0 & 190783.0 & 16091.57 \\
             pri-hum-\underline{law}             & 0.0 & 99250.0 & 6847.99 \\
             val-hum-\underline{law}             & 0.0 & 428274.0 & 26103.03 \\
             bar-hum-\underline{cha}             & 0.0 & 164446.0 & 18228.11 \\
             mon-hum-\underline{cha}             & 0.0 & 223997.0 & 18353.30 \\
             pri-hum-\underline{cha}             & 0.0 & 58367.0 & 8262.56 \\
             ran-hum-\underline{cha}             & 3.0 & 62599.0 & 8378.50 \\
             wiz-hum-\underline{cha}             & 0.0 & 55185.0 & 5316.82 \\
            \bottomrule
		\end{tabular}
        \end{adjustbox}
    \end{small}
    \end{center}
    \vskip -0.1in
\end{table}

\newpage
\section{Hyperparameters}
\label{appendix:hyperparams}

For all algorithms, hyperparameters have been reused from previous work whenever possible. For BC, CQL, and IQL reference values, see Appendix I.4 in the \citet{hambrodungeons}. For AWAC, hyperparameters from IQL were reused due to the very similar policy updating scheme. For REM, hyperparameters were taken from the original work (see \citet{agarwal2020optimistic}).

As in \citet{hambrodungeons}, and in contrast to the original CQL implementation, we multiply the TD loss by the $\alpha$ coefficient instead of the CQL loss, as we observed better results with such a scheme. We performed a search for $\alpha$ $\in [0.0001, 0.0005, 0.001, 0.01, 0.05, 0.1, 0.5, 1.0]$ with best value $\alpha = 0.0001$.

\begin{table}[h]
    \caption{BC hyperparameters.}
    \label{app:bc-hps}
    \vskip 0.1in
    \begin{center}
		\begin{tabular}{l|l}
			\toprule
            \textbf{Parameter} & \textbf{Value} \\
            \midrule
            optimizer                         & AdamW~\citep{kingma2014adam, loshchilov2017decoupled} \\
            training iterations               & 500000 \\
            batch size                        & 64 \\
            sequence length                   & 16 \\
            learning rate                     & 3e-4 \\
            weight decay                      & 0.0 \\
            state encoder                     & Chaotic-Dwarven-GPT-5\citep{hambro2022insights, hambrodungeons} \\
            LSTM hidden dim                   & 2048 \\
            LSTM layers                       & 2 \\
            LSTM dropout                      & 0.0 \\
            use previous action               & True \\
            \bottomrule
		\end{tabular}
    \end{center}
    \vskip -0.2in
\end{table}

\begin{table}[h]
    \caption{CQL hyperparameters. Note that in our implementation, the $\alpha$ coefficient multiplies the TD loss.}
    \label{app:cql-hps}
    \vskip 0.1in
    \begin{center}
		\begin{tabular}{l|l}
			\toprule
            \textbf{Parameter} & \textbf{Value} \\
            \midrule
            optimizer                         & AdamW~\citep{kingma2014adam, loshchilov2017decoupled} \\
            training iterations               & 500000 \\
            batch size                        & 64 \\
            sequence length                   & 16 \\
            learning rate                     & 3e-4 \\
            weight decay                      & 0.0 \\
            state encoder                     & Chaotic-Dwarven-GPT-5\citep{hambro2022insights, hambrodungeons} \\
            LSTM hidden dim                   & 2048 \\
            LSTM layers                       & 2 \\
            LSTM dropout                      & 0.0 \\
            use previous action               & True \\
            tau ($\tau$)                      & 5e-3 \\
            gamma ($\gamma$)                  & 0.999 \\
            reward clip range                 & [-10.0, 10.0] \\     
            alpha ($\alpha$)                  & 1e-4 \\
        \bottomrule
		\end{tabular}
    \end{center}
    \vskip -0.2in
\end{table}

\begin{table}[h]
    \caption{IQL hyperparameters.}
    \label{app:iql-hps}
    \vskip 0.1in
    \begin{center}
		\begin{tabular}{l|l}
			\toprule
            \textbf{Parameter} & \textbf{Value} \\
            \midrule
            optimizer                         & AdamW~\citep{kingma2014adam, loshchilov2017decoupled} \\
            training iterations               & 500000 \\
            batch size                        & 64 \\
            sequence length                   & 16 \\
            learning rate                     & 3e-4 \\
            weight decay                      & 0.0 \\
            state encoder                     & Chaotic-Dwarven-GPT-5\citep{hambro2022insights, hambrodungeons} \\
            LSTM hidden dim                   & 2048 \\
            LSTM layers                       & 2 \\
            LSTM dropout                      & 0.0 \\
            use previous action               & True \\
            tau ($\tau$)                      & 5e-3 \\
            gamma ($\gamma$)                  & 0.999 \\
            reward clip range                 & [-10.0, 10.0] \\ 
            expectile                         & 0.8 \\
            temperature                       & 1.0 \\
            advantage clip max                & 100 \\
            \bottomrule
	\end{tabular}
    \end{center}
    \vskip -0.2in
\end{table}

\begin{table}[h]
    \caption{AWAC hyperparameters.}
    \label{app:awac-hps}
    \vskip 0.1in
    \begin{center}
		\begin{tabular}{l|l}
			\toprule
            \textbf{Parameter} & \textbf{Value} \\
            \midrule
            optimizer                         & AdamW~\citep{kingma2014adam, loshchilov2017decoupled} \\
            training iterations               & 500000 \\
            batch size                        & 64 \\
            sequence length                   & 16 \\
            learning rate                     & 3e-4 \\
            weight decay                      & 0.0 \\
            state encoder                     & Chaotic-Dwarven-GPT-5\citep{hambro2022insights, hambrodungeons} \\
            LSTM hidden dim                   & 2048 \\
            LSTM layers                       & 2 \\
            LSTM dropout                      & 0.0 \\
            use previous action               & True \\
            tau ($\tau$)                      & 5e-3 \\
            gamma ($\gamma$)                  & 0.999 \\
            reward clip range                 & [-10.0, 10.0] \\
            temperature                       & 1.0 \\
            advantage clip max                & 100 \\
            \bottomrule
		\end{tabular}
    \end{center}
    \vskip -0.2in
\end{table}

\begin{table}[H]
    \caption{REM hyperparameters.}
    \label{app:rem-hps}
    \vskip 0.1in
    \begin{center}
		\begin{tabular}{l|l}
			\toprule
            \textbf{Parameter} & \textbf{Value} \\
            \midrule
            optimizer                         & AdamW~\citep{kingma2014adam, loshchilov2017decoupled} \\
            training iterations               & 500000 \\
            batch size                        & 64 \\
            sequence length                   & 16 \\
            learning rate                     & 3e-4 \\
            weight decay                      & 0.0 \\
            state encoder                     & Chaotic-Dwarven-GPT-5\citep{hambro2022insights, hambrodungeons} \\
            LSTM hidden dim                   & 2048 \\
            LSTM layers                       & 2 \\
            LSTM dropout                      & 0.0 \\
            use previous action               & True \\
            tau ($\tau$)                      & 5e-3 \\
            gamma ($\gamma$)                  & 0.999 \\
            reward clip range                 & [-10.0, 10.0] \\
            ensemble heads                    & 200.0 \\
            \bottomrule
		\end{tabular}
    \end{center}
    \vskip -0.2in
\end{table}

\begin{table}[h]
    \caption{The effect of CQL policy constraint strength on the performance. Results, which are averaged across 3 seeds, are sorted based on the final unnormalized game score.}
    \label{app:cql-alphas}
    \vskip 0.1in
    \begin{center}
    \begin{tabular}{l|l}
        \toprule
        \textbf{Alpha $(\alpha)$} & \textbf{Return} \\
        \midrule
          0.0001 &    526.72  $\pm$ 71.37   \\
          0.0005 &    396.50  $\pm$ 39.35   \\
          0.001  &    395.56  $\pm$ 139.92  \\
          0.05   &    226.55  $\pm$ 196.45  \\
          0.01   &     32.42  $\pm$ 43.69   \\
          0.1    &     14.04  $\pm$ 10.29   \\
          0.5    &      0.00  $\pm$ 0.00    \\
          1.0    &      0.59  $\pm$ 0.45    \\
        \bottomrule
    \end{tabular}
    \end{center}
    \vskip -0.2in
\end{table}

\newpage
\section{Results per Benchmark Categories}
\label{appendix:results-stratified}

In this section, we report the results stratified by the introduced categories. If one is willing to inspect specific datasets, we organized all training logs into Weights\&Biases public reports, found at \url{https://wandb.ai/tlab/NetHack/reports}. 

Note that one can find all the evaluation scores (for more than one checkpoint) within the runs and use them for any evaluation tools of interest. Also, we provide convenient scripts for constructing RLiable \citep{agarwal2021deep} graphs based on the provided runs that can be configured for one's purposes as well (see \url{https://github.com/corl-team/katakomba/tree/main/scripts/rliable_report.py}).

\begin{figure}[ht]
\centering
\captionsetup{justification=centering}
     \centering
         \begin{subfigure}[b]{0.98\textwidth}
         \centering
         \includegraphics[width=1.0\textwidth]{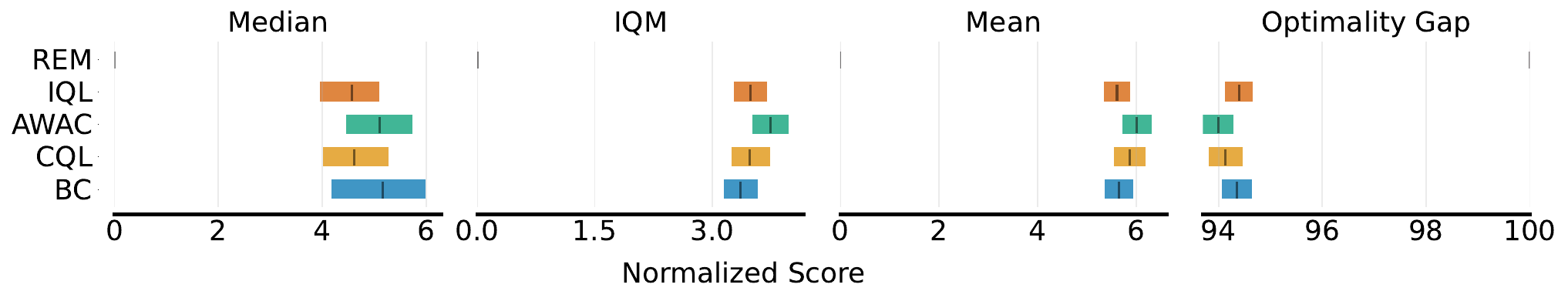}
         \caption{Bootstrapped point estimates.}
         \label{appendix:fig:ns:base:point-estimates}
        \end{subfigure}
        \begin{subfigure}[b]{0.49\textwidth}
         \centering
         \includegraphics[width=1.0\textwidth]{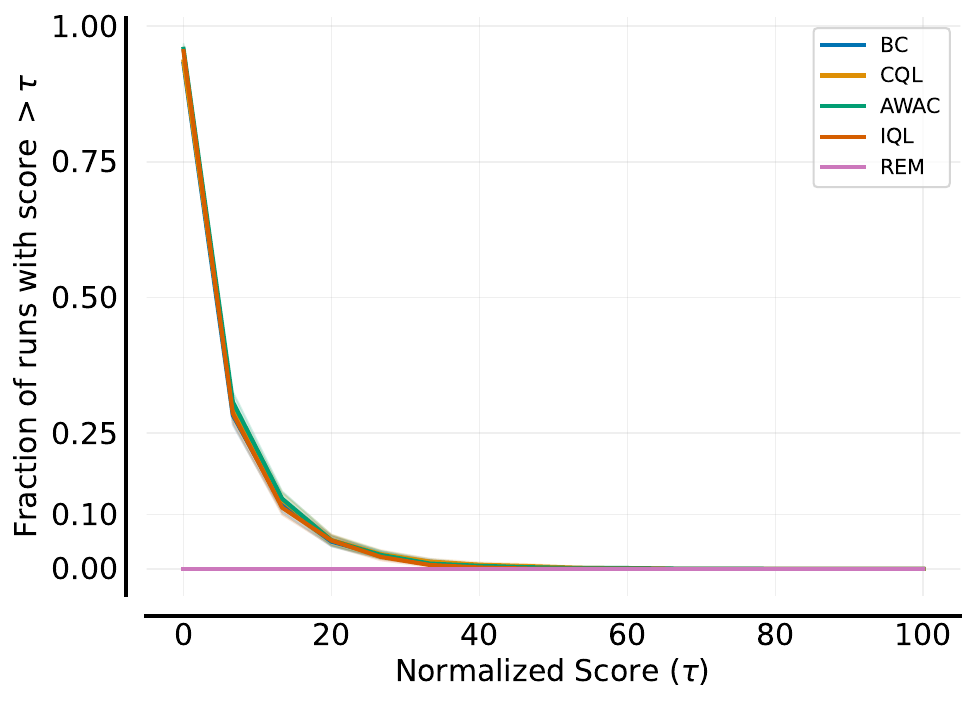}
         \caption{Performance profiles.}
         \label{appendix:fig:ns:base:performance-profiles}
        \end{subfigure}
        \begin{subfigure}[b]{0.49\textwidth}
         \centering
         \includegraphics[width=1.0\textwidth]{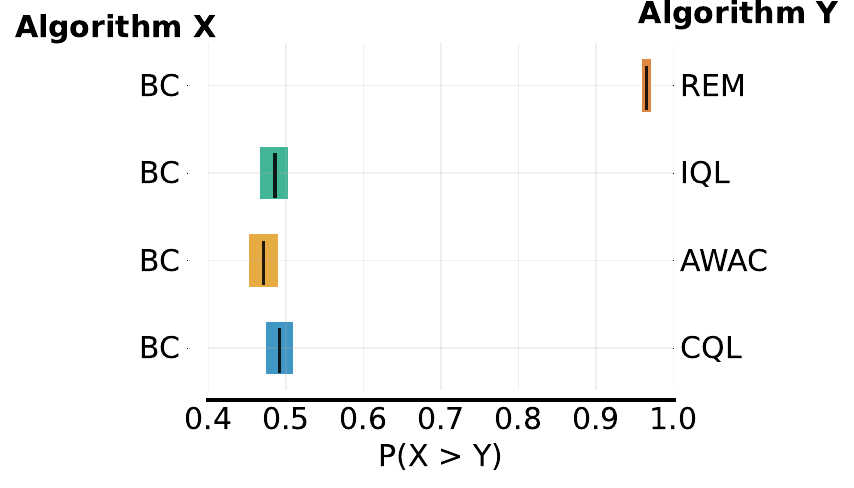}
         \caption{Probability of improvement of BC to other algorithms.}
         \label{appendix:fig:ns:base:probability-improvement}
        \end{subfigure}
        \caption{{Normalized performance under the Katakomba benchmark for \underline{Base} datasets. Each algorithm was run for three seeds and evaluated over 50 episodes resulting in 1950 points for constructing these graphs.}}
        \label{appendix:fig:base:ns}
\end{figure}

\begin{figure}[ht]
\centering
\captionsetup{justification=centering}
     \centering
         \begin{subfigure}[b]{0.98\textwidth}
         \centering
         \includegraphics[width=1.0\textwidth]{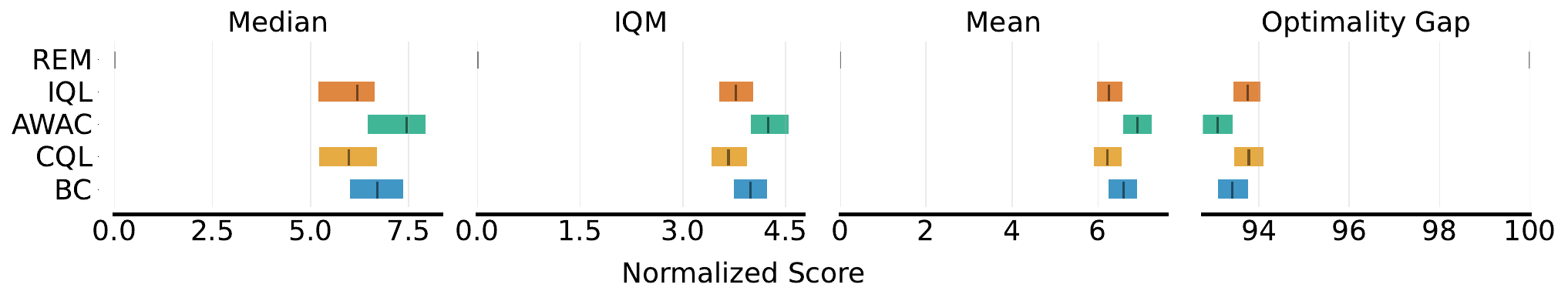}
         \caption{Bootstrapped point estimates.}
         \label{appendix:fig:ns:extended:point-estimates}
        \end{subfigure}
        \begin{subfigure}[b]{0.49\textwidth}
         \centering
         \includegraphics[width=1.0\textwidth]{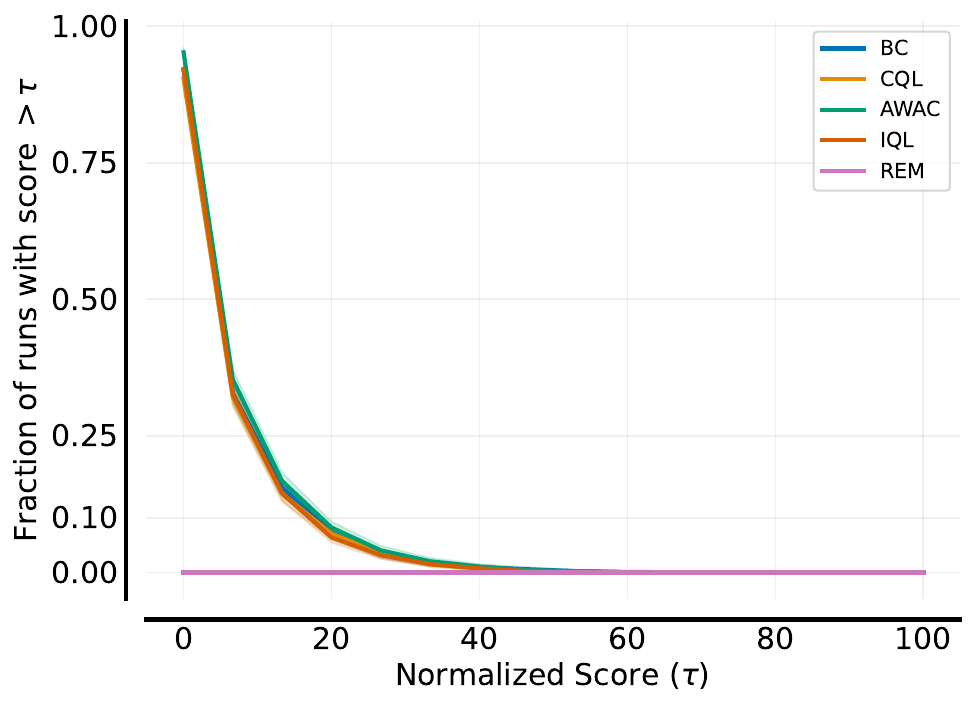}
         \caption{Performance profiles.}
         \label{appendix:fig:ns:extended:performance-profiles}
        \end{subfigure}
        \begin{subfigure}[b]{0.49\textwidth}
         \centering
         \includegraphics[width=1.0\textwidth]{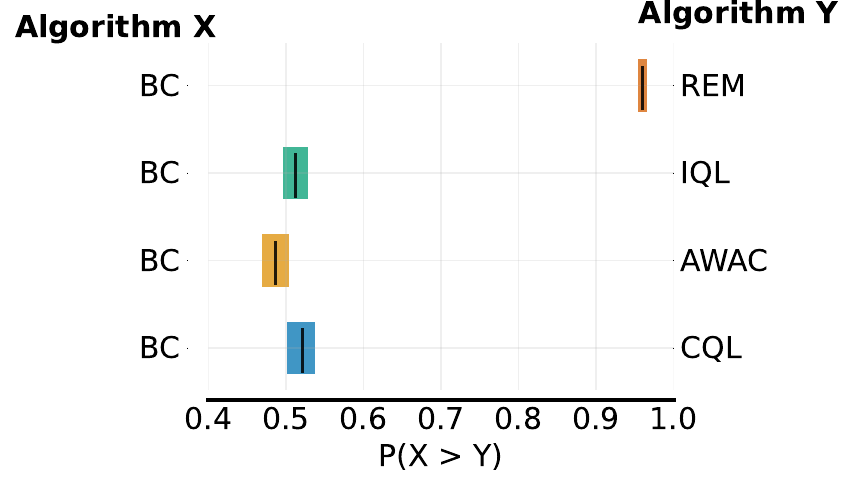}
         \caption{Probability of improvement of BC to other algorithms.}
         \label{appendix:fig:ns:extended:probability-improvement}
        \end{subfigure}
        \caption{{Normalized performance under the Katakomba benchmark for \underline{Extended} datasets. Each algorithm was run for three seeds and evaluated over 50 episodes resulting in 2250 points for constructing these graphs.}}
        \label{appendix:fig:extended:ns}
\end{figure}

\begin{figure}[ht]
\centering
\captionsetup{justification=centering}
     \centering
         \begin{subfigure}[b]{0.98\textwidth}
         \centering
         \includegraphics[width=1.0\textwidth]{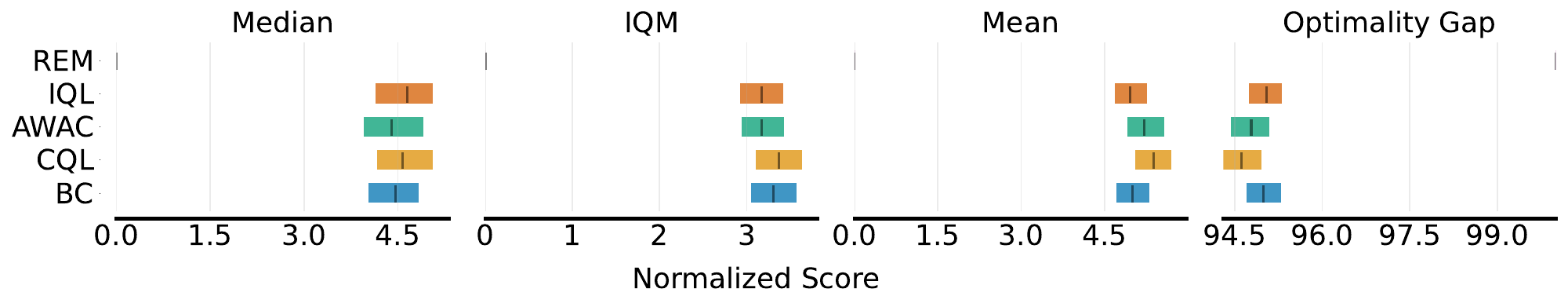}
         \caption{Bootstrapped point estimates.}
         \label{appendix:fig:ns:complete:point-estimates}
        \end{subfigure}
        \begin{subfigure}[b]{0.49\textwidth}
         \centering
         \includegraphics[width=1.0\textwidth]{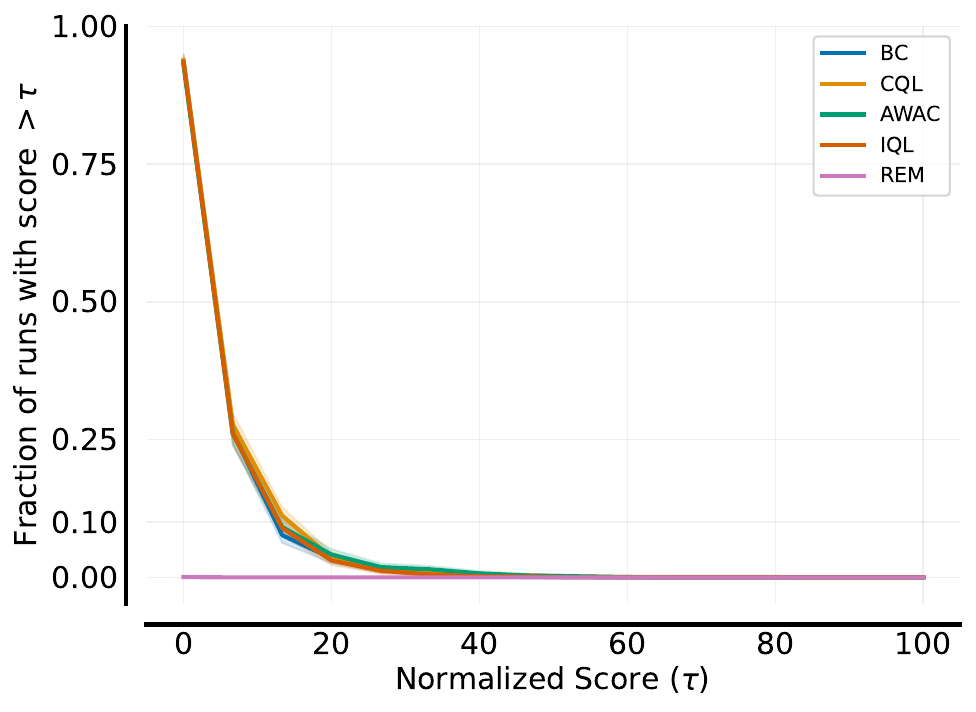}
         \caption{Performance profiles.}
         \label{appendix:fig:ns:complete:performance-profiles}
        \end{subfigure}
        \begin{subfigure}[b]{0.49\textwidth}
         \centering
         \includegraphics[width=1.0\textwidth]{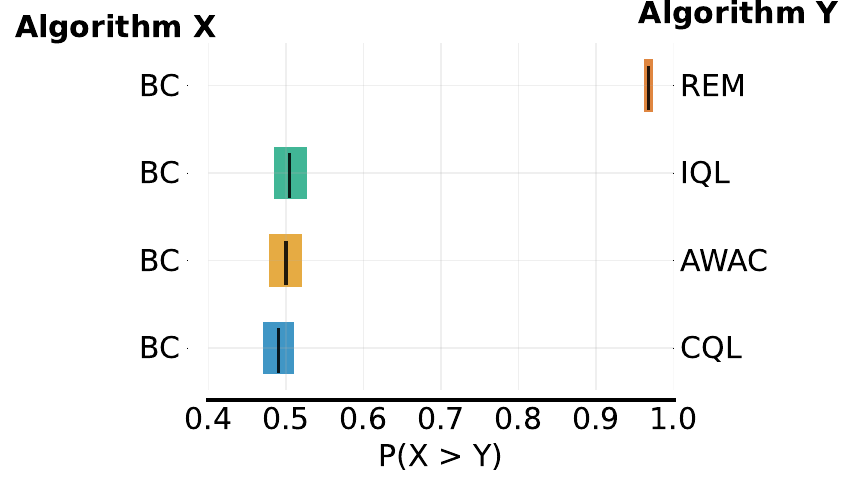}
         \caption{Probability of improvement of BC to other algorithms.}
         \label{appendix:fig:ns:complete:probability-improvement}
        \end{subfigure}
        \caption{{Normalized performance under the Katakomba benchmark for \underline{Complete} datasets. Each algorithm was run for three seeds and evaluated over 50 episodes resulting in 1500 points for constructing these graphs.}}
        \label{appendix:fig:complete:ns}
\end{figure}

\begin{figure}[ht]
\centering
\captionsetup{justification=centering}
     \centering
         \begin{subfigure}[b]{0.98\textwidth}
         \centering
         \includegraphics[width=1.0\textwidth]{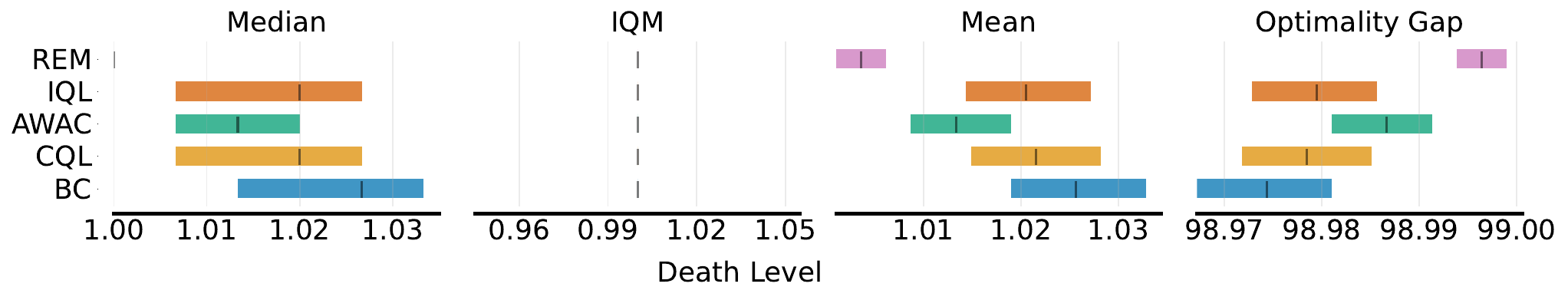}
         \caption{Bootstrapped point estimates.}
         \label{appendix:fig:depth:base:point-estimates}
        \end{subfigure}
        \begin{subfigure}[b]{0.49\textwidth}
         \centering
         \includegraphics[width=1.0\textwidth]{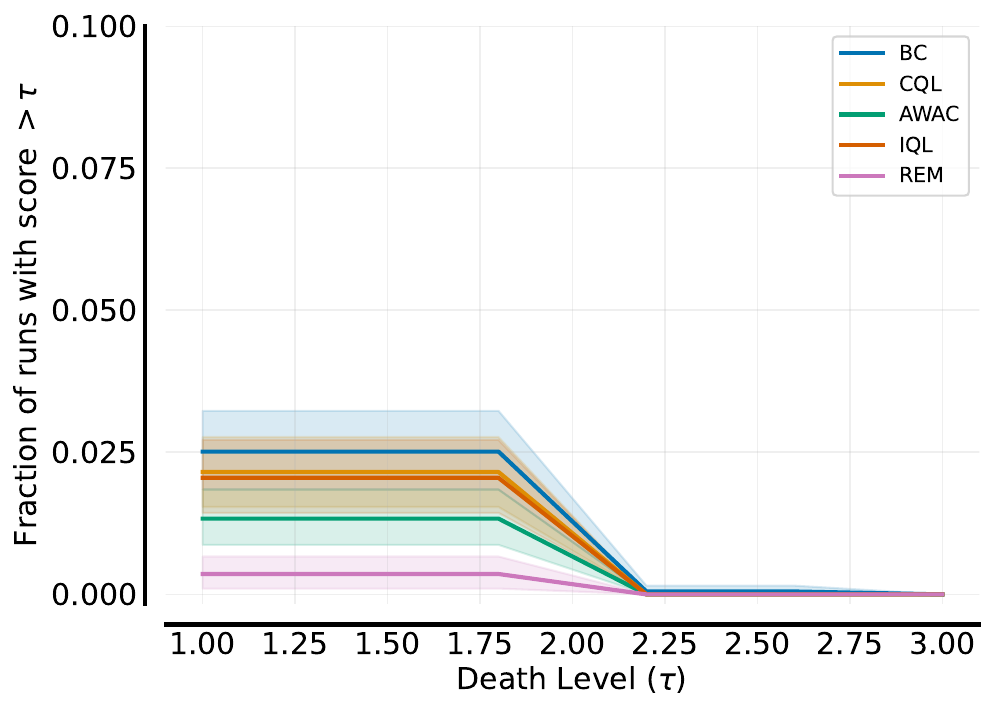}
         \caption{Performance profiles.}
         \label{appendix:fig:depth:base:performance-profiles}
        \end{subfigure}
        \begin{subfigure}[b]{0.49\textwidth}
         \centering
         \includegraphics[width=1.0\textwidth]{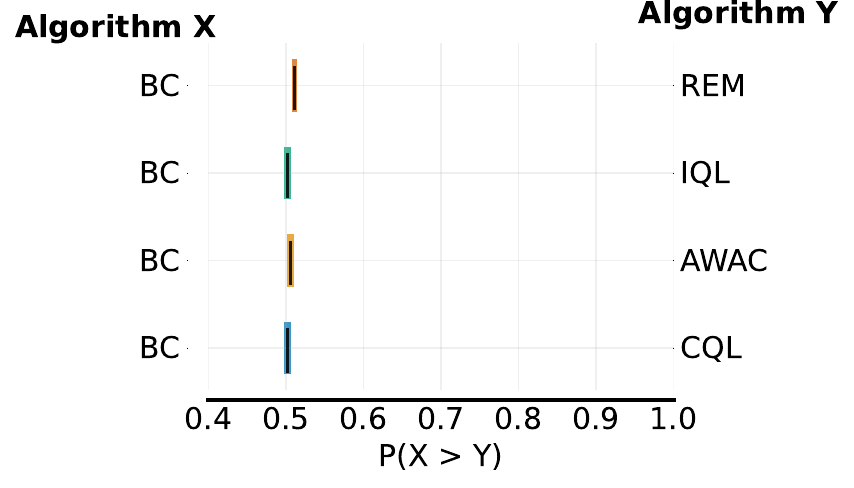}
         \caption{Probability of improvement of BC to other algorithms.}
         \label{appendix:fig:depth:base:probability-improvement}
        \end{subfigure}
        \caption{{Death levels under the Katakomba benchmark for \underline{Base} datasets. Each algorithm was run for three seeds and evaluated over 50 episodes resulting in 1950 points for constructing these graphs.}}
        \label{appendix:fig:base:depth}
\end{figure}

\begin{figure}[ht]
\centering
\captionsetup{justification=centering}
     \centering
         \begin{subfigure}[b]{0.98\textwidth}
         \centering
         \includegraphics[width=1.0\textwidth]{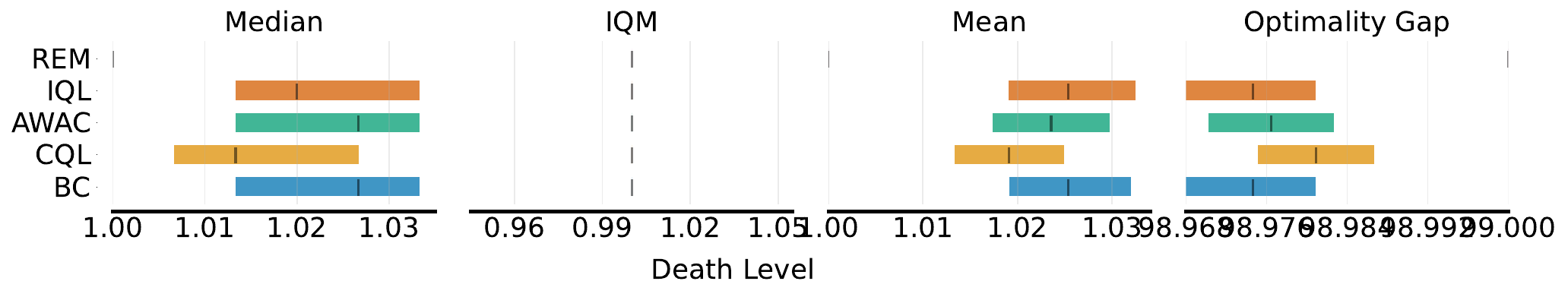}
         \caption{Bootstrapped point estimates.}
         \label{appendix:fig:depth:extended:point-estimates}
        \end{subfigure}
        \begin{subfigure}[b]{0.49\textwidth}
         \centering
         \includegraphics[width=1.0\textwidth]{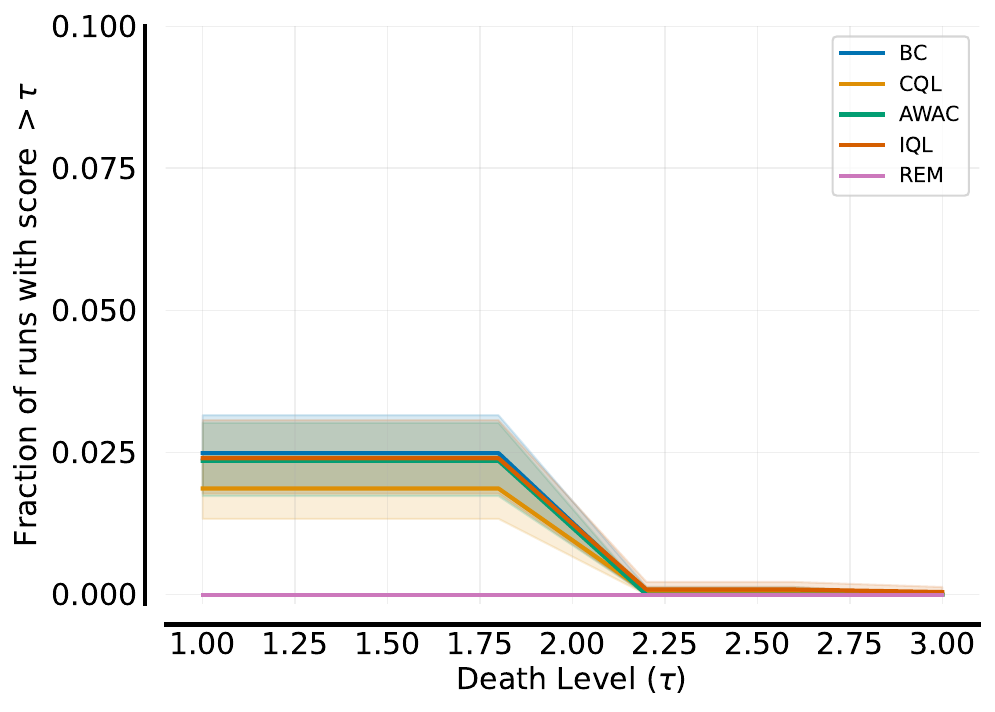}
         \caption{Performance profiles.}
         \label{appendix:fig:depth:extended:performance-profiles}
        \end{subfigure}
        \begin{subfigure}[b]{0.49\textwidth}
         \centering
         \includegraphics[width=1.0\textwidth]{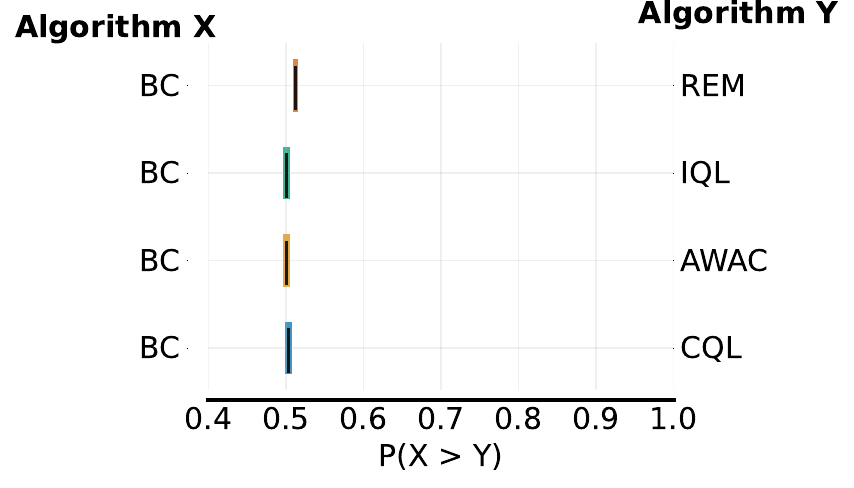}
         \caption{Probability of improvement of BC to other algorithms.}
         \label{appendix:fig:depth:extended:probability-improvement}
        \end{subfigure}
        \caption{{Death level under the Katakomba benchmark for \underline{Extended} datasets. Each algorithm was run for three seeds and evaluated over 50 episodes resulting in 2250 points for constructing these graphs.}}
        \label{appendix:fig:extended:depth}
\end{figure}

\begin{figure}[ht]
\centering
\captionsetup{justification=centering}
     \centering
         \begin{subfigure}[b]{0.98\textwidth}
         \centering
         \includegraphics[width=1.0\textwidth]{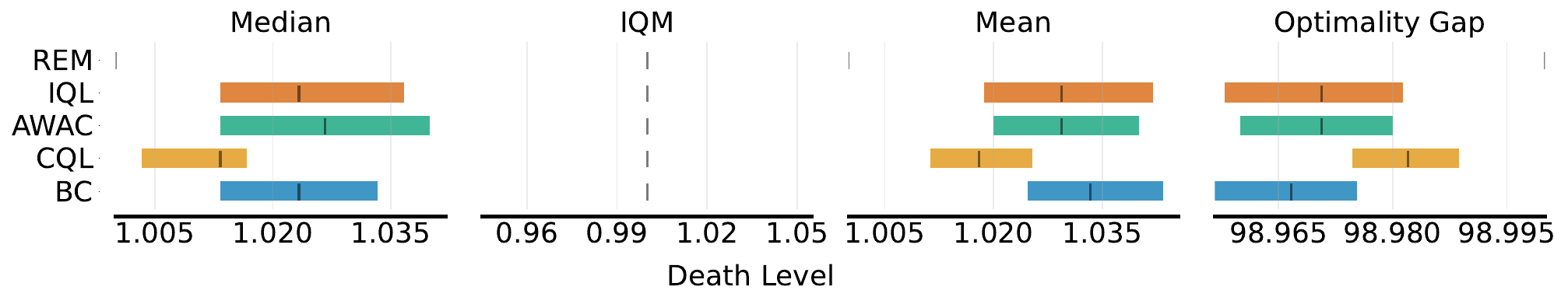}
         \caption{Bootstrapped point estimates.}
         \label{appendix:fig:depth:complete:point-estimates}
        \end{subfigure}
        \begin{subfigure}[b]{0.49\textwidth}
         \centering
         \includegraphics[width=1.0\textwidth]{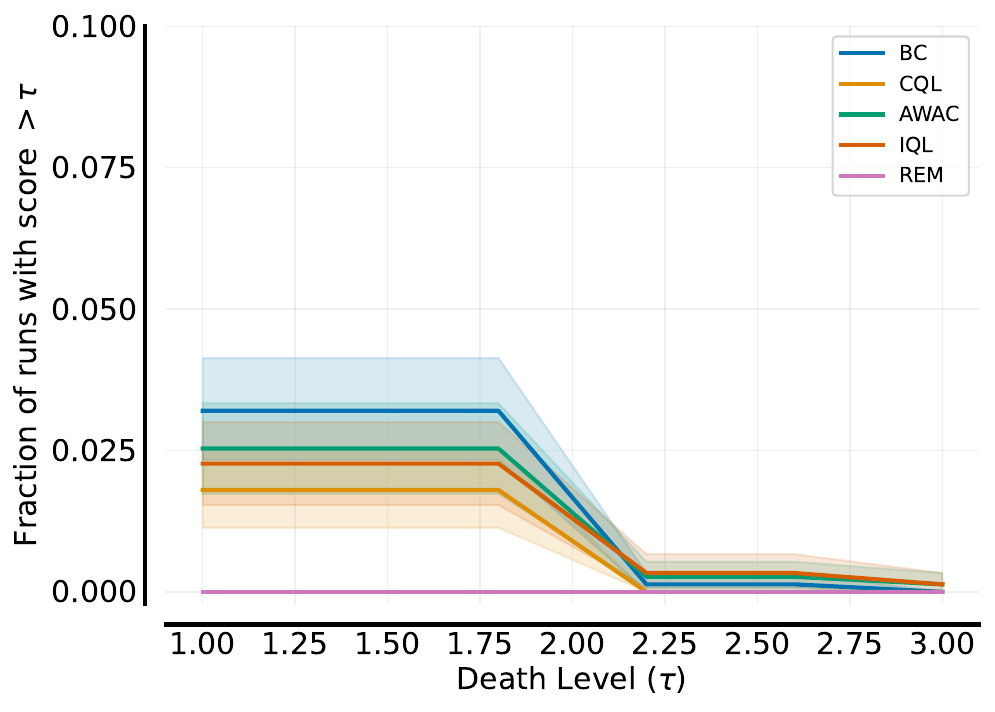}
         \caption{Performance profiles.}
         \label{appendix:fig:depth:complete:performance-profiles}
        \end{subfigure}
        \begin{subfigure}[b]{0.49\textwidth}
         \centering
         \includegraphics[width=1.0\textwidth]{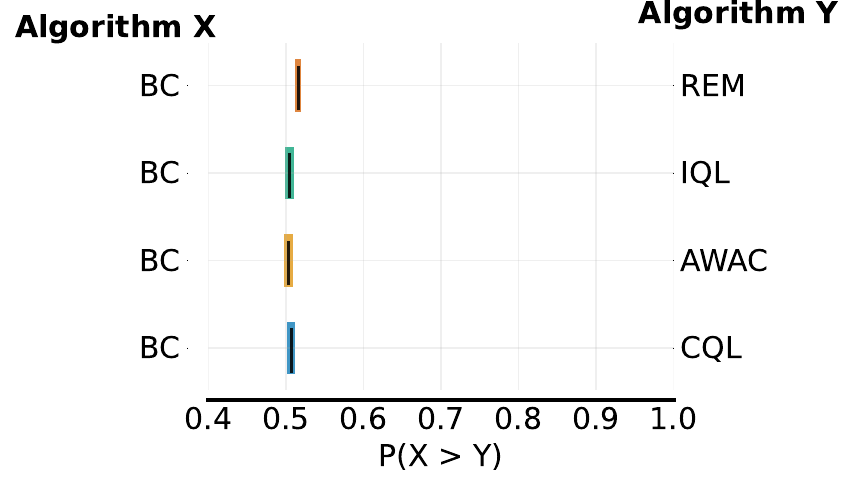}
         \caption{Probability of improvement of BC to other algorithms.}
         \label{appendix:fig:depth:complete:probability-improvement}
        \end{subfigure}
        \caption{{Death levels under the Katakomba benchmark for \underline{Complete} datasets. Each algorithm was run for three seeds and evaluated over 50 episodes resulting in 1500 points for constructing these graphs.}}
        \label{appendix:fig:complete:depth}
\end{figure}

\begin{figure}[ht]
\centering
\captionsetup{justification=centering}
     \centering
         \begin{subfigure}[b]{0.98\textwidth}
         \centering
         \includegraphics[width=1.0\textwidth]{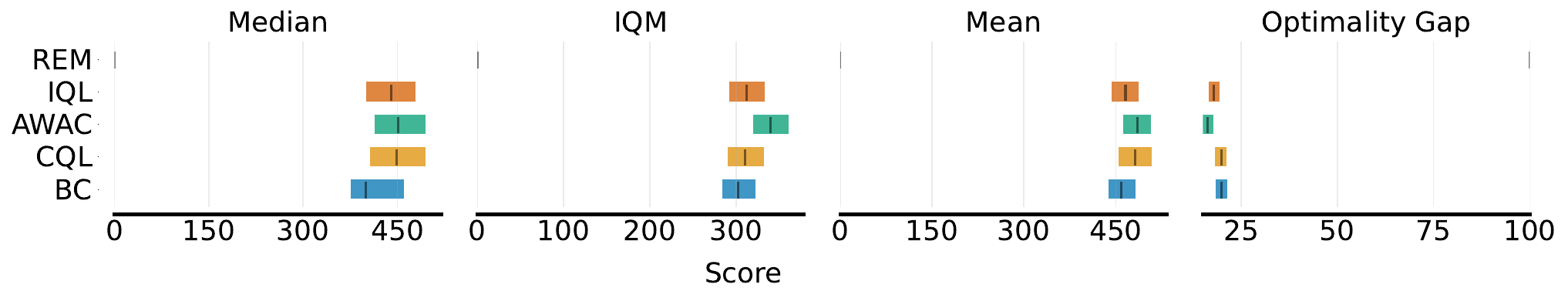}
         \caption{Bootstrapped point estimates.}
         \label{appendix:fig:scores:base:point-estimates}
        \end{subfigure}
        \begin{subfigure}[b]{0.49\textwidth}
         \centering
         \includegraphics[width=1.0\textwidth]{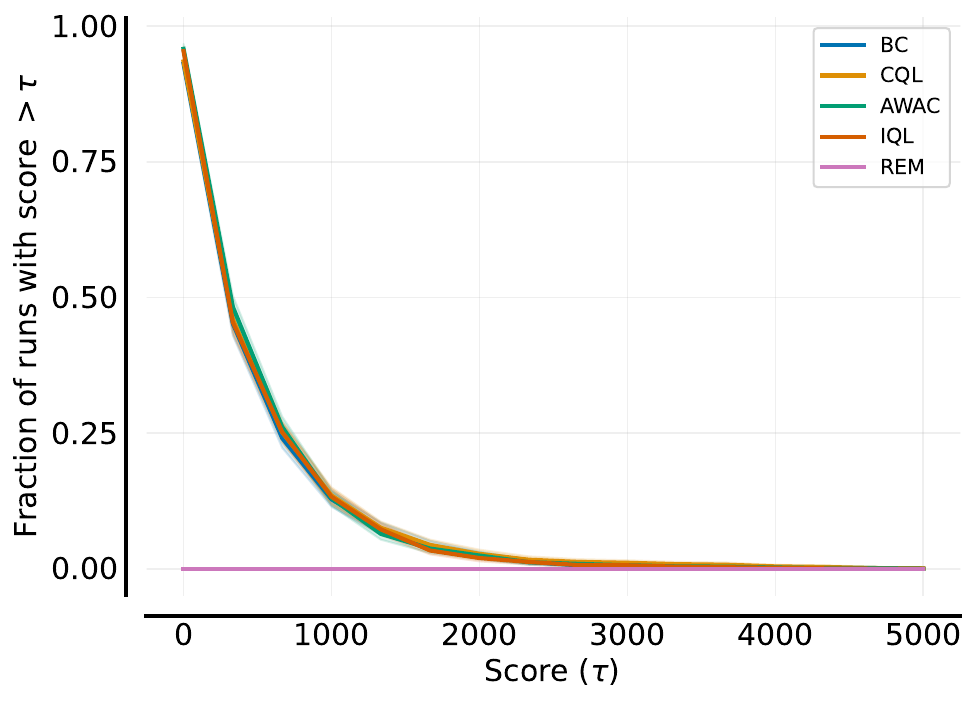}
         \caption{Performance profiles.}
         \label{appendix:fig:scores:base:performance-profiles}
        \end{subfigure}
        \begin{subfigure}[b]{0.49\textwidth}
         \centering
         \includegraphics[width=1.0\textwidth]{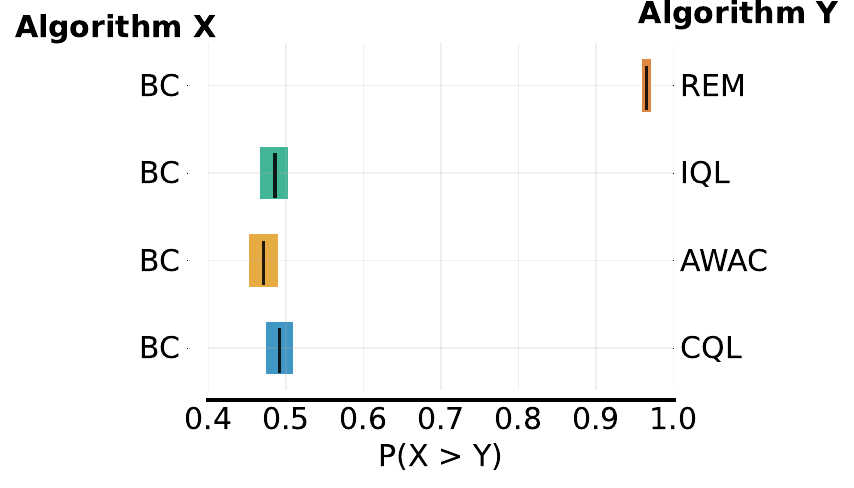}
         \caption{Probability of improvement of BC to other algorithms.}
         \label{appendix:fig:scores:base:probability-improvement}
        \end{subfigure}
        \caption{{Unnormalized in-game score under the Katakomba benchmark for \underline{Base} datasets. Each algorithm was run for three seeds and evaluated over 50 episodes resulting in 1950 points for constructing these graphs.}}
        \label{appendix:fig:base:scores}
\end{figure}

\begin{figure}[ht]
\centering
\captionsetup{justification=centering}
     \centering
         \begin{subfigure}[b]{0.98\textwidth}
         \centering
         \includegraphics[width=1.0\textwidth]{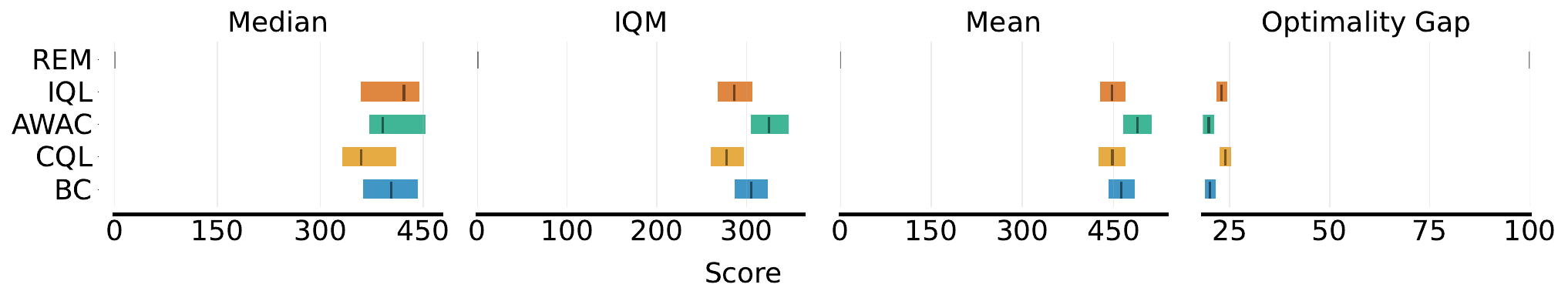}
         \caption{Bootstrapped point estimates.}
         \label{appendix:fig:scores:extended:point-estimates}
        \end{subfigure}
        \begin{subfigure}[b]{0.49\textwidth}
         \centering
         \includegraphics[width=1.0\textwidth]{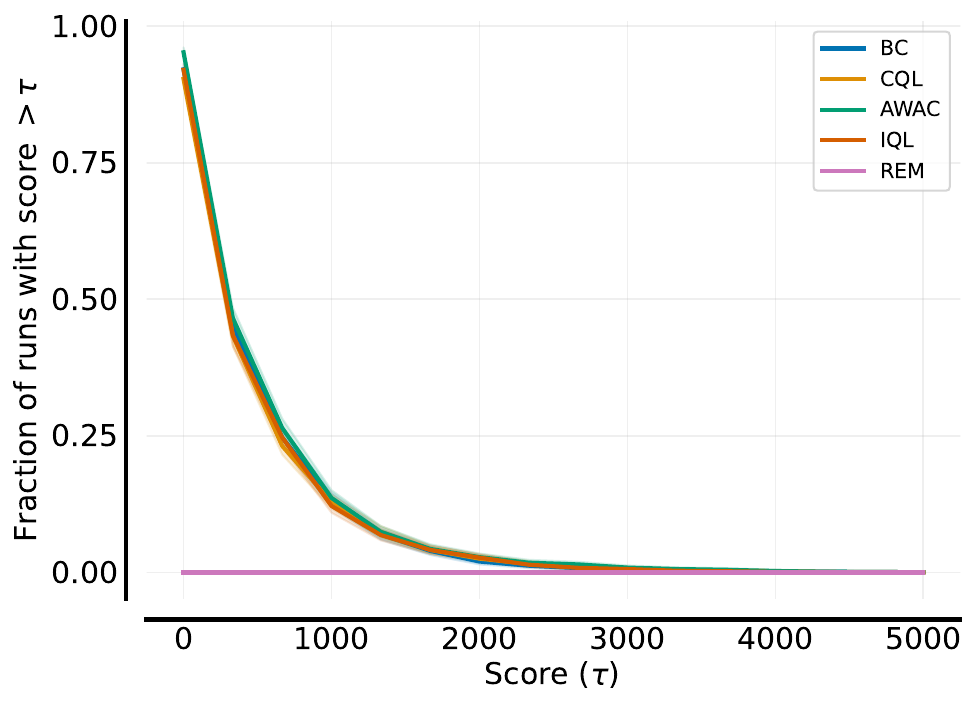}
         \caption{Performance profiles.}
         \label{appendix:fig:scores:extended:performance-profiles}
        \end{subfigure}
        \begin{subfigure}[b]{0.49\textwidth}
         \centering
         \includegraphics[width=1.0\textwidth]{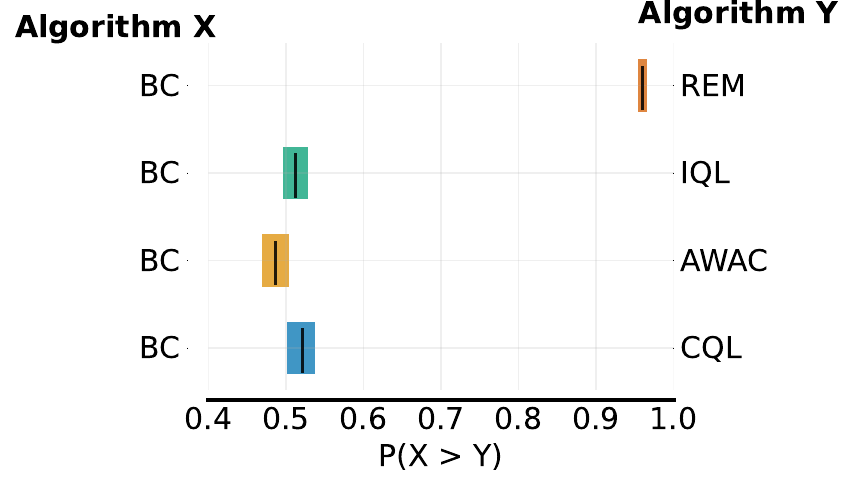}
         \caption{Probability of improvement of BC to other algorithms.}
         \label{appendix:fig:scores:extended:probability-improvement}
        \end{subfigure}
        \caption{{Unnormalized in-game score under the Katakomba benchmark for \underline{Extended} datasets. Each algorithm was run for three seeds and evaluated over 50 episodes resulting in 2250 points for constructing these graphs.}}
        \label{appendix:fig:extended:scores}
\end{figure}

\begin{figure}[ht]
\centering
\captionsetup{justification=centering}
     \centering
         \begin{subfigure}[b]{0.98\textwidth}
         \centering
         \includegraphics[width=1.0\textwidth]{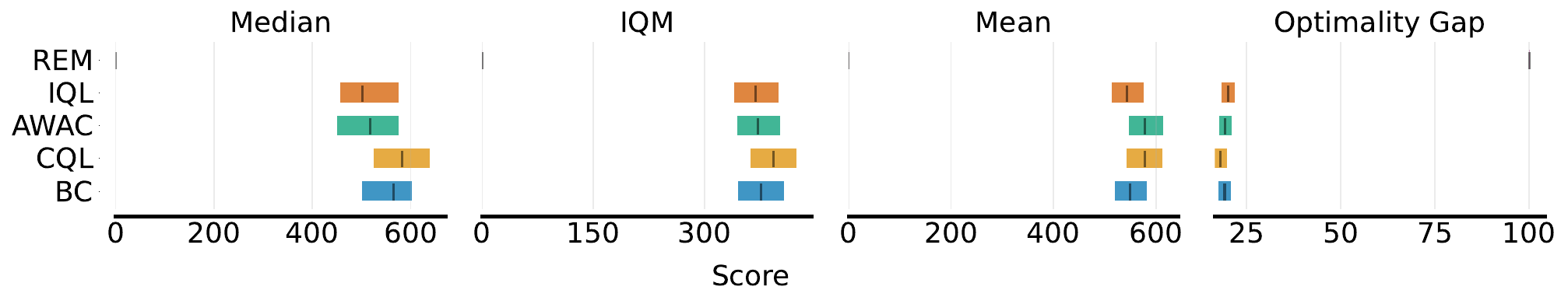}
         \caption{Bootstrapped point estimates.}
         \label{appendix:fig:scores:complete:point-estimates}
        \end{subfigure}
        \begin{subfigure}[b]{0.49\textwidth}
         \centering
         \includegraphics[width=1.0\textwidth]{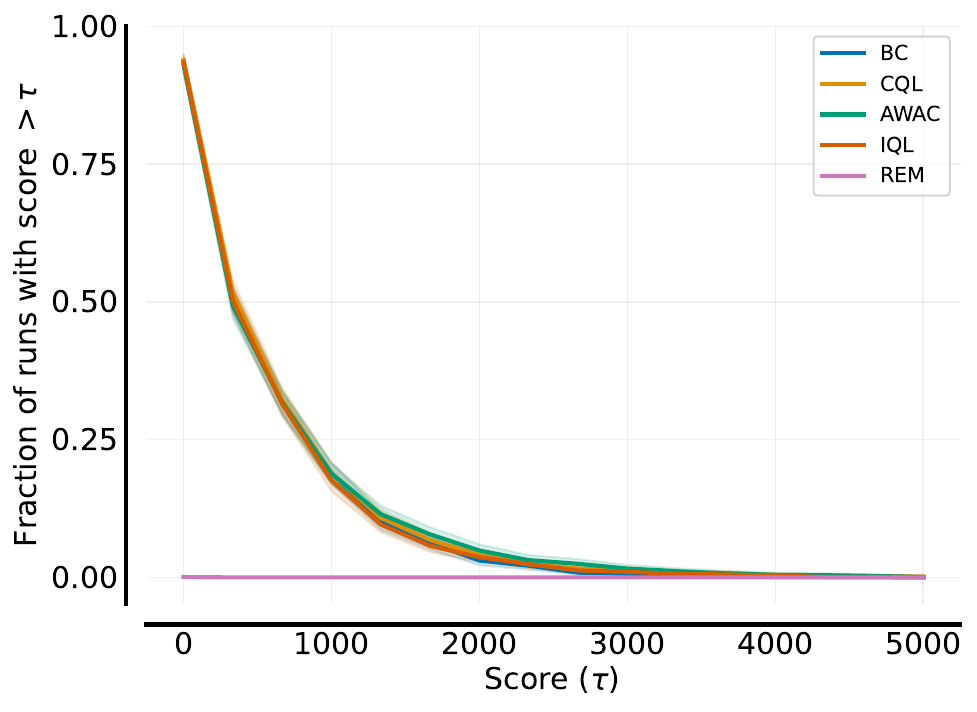}
         \caption{Performance profiles.}
         \label{appendix:fig:scores:complete:performance-profiles}
        \end{subfigure}
        \begin{subfigure}[b]{0.49\textwidth}
         \centering
         \includegraphics[width=1.0\textwidth]{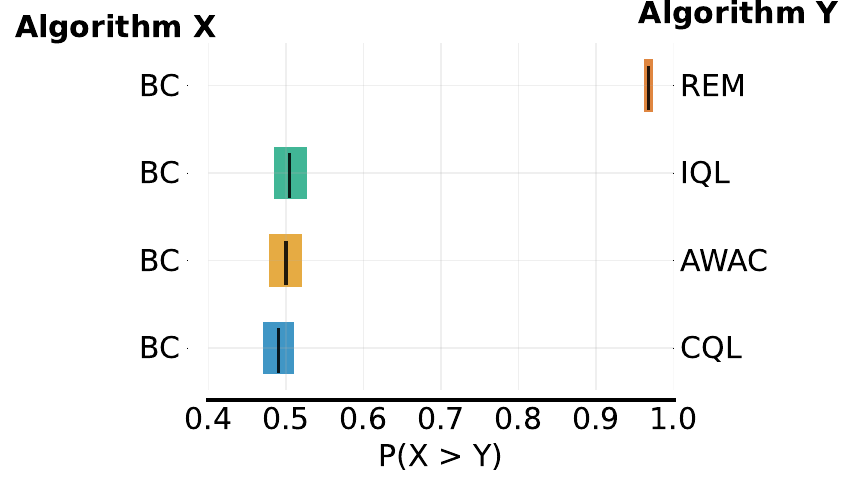}
         \caption{Probability of improvement of BC to other algorithms.}
         \label{appendix:fig:scores:complete:probability-improvement}
        \end{subfigure}
        \caption{{Unnormalized in-game score under the Katakomba benchmark for \underline{Complete} datasets. Each algorithm was run for three seeds and evaluated over 50 episodes resulting in 1500 points for constructing these graphs.}}
        \label{appendix:fig:complete:scores}
\end{figure}
\end{document}